\documentclass{article}
\usepackage[utf8]{inputenc}
\usepackage[T1]{fontenc}
\usepackage{siunitx}
\usepackage{ifthen}%
\usepackage{mfirstuc}
\usepackage{xkeyval}%
\usepackage{xfor}%
\usepackage{amsgen}%
\usepackage{etoolbox}%
\usepackage{datatool-base}%
\usepackage{listings}
\usepackage{lscape}
\usepackage{cite}
\usepackage[%
  xindy,%
  style=long,%
  toc=true,%
  nonumberlist,%
  acronym,%
  acronymlists={abbreviations},%
  hyperfirst=true% link first use of abbreviation
  ]{glossaries}%

\newacronym{AI}{AI}{artificial intelligence}
\newacronym{AUC}{AUC}{area under the curve}
\newacronym{EHS}{EHS}{Early Help Services}
\newacronym{FN}{FN}{false negative}
\newacronym{FNR}{FNR}{false negative rate}
\newacronym{FP}{FP}{false positive}
\newacronym{FPR}{FPR}{false positive rate}
\newacronym{GBC}{GBC}{gradient-boosting classifier}
\newacronym{LR}{LR}{logistic regression}
\newacronym{ML}{ML}{machine learning}
\newacronym{LCC}{LCC}{Leicestershire County Council}
\newacronym{LIME}{LIME}{local interpretable model-agnostic explanations}
\newacronym{RF}{RF}{random forest}
\newacronym{ROC}{ROC}{receiver operating characteristic}
\newacronym{TN}{TN}{true negative}
\newacronym{TNR}{TNR}{true negative rate}
\newacronym{TP}{TP}{true positive}
\newacronym{TPR}{TPR}{true positive rate}
\newacronym{PRU}{\textsc{pru}}{pupil referral unit}
\newacronym{SENDEHC}{\textsc{send ehc}}{special educational needs and disabilities for education, health and care}
\newacronym{SEND}{\textsc{send needs}}{special educational needs and disabilities}
\newacronym{REFERRAL}{\textsc{send referral}}{referred for special educational needs and disabilities}
\newacronym{NEET}{\textsc{neet}}{Not in Education, Employment or Training}
\newacronym{TRANSFER}{\textsc{transfer phased}}{School Transfer Phased}
\newacronym{FSM}{\textsc{fsm}}{free school meal}
\newacronym{EYF}{\textsc{eyf}}{early years funding}
\newacronym{CFWS}{CFWS}{Children and Family Wellbeing Services}
\newacronym{CV}{CV}{cross-validation}

\usepackage{subcaption}
\usepackage{multicol}
\usepackage{multirow} 
\usepackage{arxiv}
\usepackage[utf8]{inputenc} % allow utf-8 input
\usepackage[T1]{fontenc}    % use 8-bit T1 fonts
\usepackage{url}            % simple URL typesetting
\usepackage{booktabs}       % professional-quality tables
\usepackage{amsfonts}       % blackboard math symbols
\usepackage{nicefrac}       % compact symbols for 1/2, etc.
\usepackage{microtype}      % microtypography
\usepackage{lipsum}		% Can be removed after putting your text content
\usepackage{graphicx}
\usepackage{doi}

\title{Identifying Early Help Referrals for Local Authorities with Machine Learning and Bias Analysis}

\author{{\hspace{1mm}Eufr\'asio de A. Lima Neto} \\
School Computer Science and Informatics\\
De Montfort University \\
Leicester, United Kingdom\\
	\And
{\hspace{1mm}Jonathan Bailiss, Axel Finke } \\
School of Science\\
Loughborough University \\
Loughborough, United Kingdom\\
	\And
	{\hspace{1mm}Jo Miller} \\
	Leicestershire County Council\\
	Glenfield\\
	 Leicester, United Kingdom\\
	\And
	{\hspace{1mm}Georgina Cosma} \\
	School of Science\\
	Loughborough University \\
 Loughborough, United Kingdom\\
	\texttt{corresponding author: g.cosma@lboro.ac.uk} \\
}

\renewcommand{\shorttitle}{Identifying Early Help Referrals for Local Authorities with ML and Bias Analysis}

\hypersetup{
pdftitle={Identifying Early Help Referrals for Local Authorities with Machine Learning and Bias Analysis},
pdfsubject={q-bio.NC, q-bio.QM},
pdfauthor={Eufr\'asio de A. Lima Neto, Jonathan Bailiss, Axel Finke, Jo Miller, Georgina Cosma},
pdfkeywords={First keyword, Second keyword, More},
}

\begin{document}
\maketitle

\begin{abstract}
Local authorities in England, such as Leicestershire County Council (LCC), provide Early Help services that can be offered at any point in a young person's life when they experience difficulties that cannot be supported by universal services alone, such as schools. This paper investigates the utilisation of machine learning (ML) to assist experts in identifying families that may need to be referred for Early Help assessment and support. LCC provided an anonymised dataset comprising 14\,360 records of young people under the age of 18. The dataset was pre-processed, machine learning models were build, and experiments were conducted to validate and test the performance of the models. Bias mitigation techniques were applied to improve the fairness of these models. During testing, while the models demonstrated the capability to identify young people requiring intervention or early help, they also produced a significant number of false positives, especially when constructed with imbalanced data, incorrectly identifying individuals who most likely did not need an Early Help referral. This paper empirically explores the suitability of data-driven ML models for identifying young people who may require Early Help services and discusses their appropriateness and limitations for this task.
\end{abstract}

% keywords can be removed
\section{Introduction}
Local authorities in England have statutory responsibility for protecting the welfare of children and delivering children's social care. The COVID-19 pandemic has put pressure on the children's social care sector, and exacerbated existing challenges in risk and other assessments \cite{tupper2017, covid2022children}. Early Help is a service provided by Local Authorities that offers social care and support to families, including early intervention services for children, young people, or families facing challenges beyond the scope of universal services like schools or general practitioners. Early Help provides services that meet the needs of families who have lower-level support needs (e.g. not child protection) to prevent problems from escalating and entering the social care system (e.g. child protection). Furthermore, Early Help can offer children the support needed to reach their full potential; improve the quality of a child's home and family life; enable them to perform better at school and support their mental health; and can also support a child to develop strengths and skills that can prepare them for adult life.\\
\textbf{Increase of children in need.} Data collected from the most recent Child in Need Census 2022--2023 revealed that there are: 404\,310 Children in need, up 4.1\% from 2021 and up 3.9\% from 2020. This is the highest number of children in need since 2018. Furthermore, there were 650\,270 referrals, up 8.8\% from 2021 and up 1.1\% from 2020. This is the highest number of referrals since 2019. In 2022, compared with 2021 when restrictions on school attendance were in place for parts of the year due to COVID-19, referrals from schools increased, in turn driving the overall rise in referrals. The Department for Education (DfE) is collecting data on early help provision as part of the Child in Need Census 2022--2023 \cite{childinneed2022}, 2023--2024 Cescus \cite{childinneed2023}. This census data is submitted by local authorities to the DfE between early April and end of July of each year. Information about early help enables DfE to understand more about the contact that children in need have with the Early Help services that local authorities provide.\\
\textbf{Need for data-driven tools and machine learning.} The increasing numbers of children in need and referrals have highlighted the need for data-driven tools that can analyse large datasets to aid local authorities in making informed decisions for individuals at risk alleviating pressure on a chronically overstretched service. \Gls{ML}, an application of \gls{AI}, can efficiently analyse vast amounts of data from diverse sources. In the context of children's services, this capability allows for the identification of risk factors of which social workers may not have otherwise been aware, such as a family falling behind on rent payments. Such data can be combined with other relevant information, such as school attendance records, to provide a more comprehensive view of the situation. \\
The effectiveness of \gls{ML} is currently limited by the lack of transparency \cite{cutillo2020machine} in the decision-making process of \gls{ML} models \cite{Aniek2021} and the data they use \cite{liu2022trustworthy, thiebes2021trustworthy}. Therefore, it is crucial to increase transparency in \gls{ML} decision-making processes to ensure fair and equitable outcomes for individuals and communities. A \gls{ML} model may be \textit{biased} if it systematically performs better on certain socio-demographic groups \cite{mehrabi2021survey, ntoutsi2020bias}. This can occur when the model has been developed on unrepresentative, incomplete, faulty, or prejudicial data \cite{fazelpour2021algorithmic, mhasawade2021machine, nijman2022missing}. Given the potential impact of bias on individuals and society, there is growing interest among businesses, governments, and organizations in tools and techniques for detecting, visualising, and mitigating bias in \gls{ML}. These tools have gained popularity in addressing bias-related issues and are increasingly recognized as important solutions to promoting fairness \cite{wisniewski2021fairmodels, kamiran2012data, feldman2015certifying, calmon2017, zhang2018mitigating,
baniecki2021dalex, gupta2021individual, konstantinov2022fairness}. The use of \gls{ML} in social care presents a topic of discussion that involves technical and ethical considerations \cite{leslie2020ethics}. For example, if used responsibly and fairly, these models have the potential to assist in protecting young people \cite{glaberson2019coding}, particularly when combined with successful early intervention programs such as the Early Start program developed in New Zealand \cite{gillingham2016predictive}. Responsible use of \gls{ML} models has the potential to enhance the usefulness of risk assessment tools in child welfare \cite{walsh2020exploring}.\\
This paper evaluates the suitability of \gls{ML} models for identifying young people who may require \gls{EHS} and applies methods for identifying and mitigating bias. For the purposes of this work, \gls{LCC}'s locality triage will be categorised as (1) \textsc{eh support}: Early Help support: the most intense type of intervention; (2) \textsc{some action}: referral to less intensive services such as group activities or schemes that run during the school holidays, or to external services; (3) \textsc{no action}: additional support is not currently required. Specifically, the contributions of this paper are: (a) \gls{ML} models were implemented and their performance was evaluated across different validation and test sets; and (b) bias analysis was conducted and mitigation algorithms were applied to reduce bias in the \gls{ML} models. This study revealed that certain educational indicators such as fixed-term exclusion and free school meals may predict the need for \gls{EHS}.
\section{Methods}

\subsection{Dataset}
The dataset contains records of young people who are under 18 years, and the dataset was provided by the \gls{LCC}. The data relates to families and individuals assessed for Early Help support between April 2019 and August 2022. The time period of data included within the features varies depending on the age of the young person with older individuals having data across a longer time-frame. The initial dataset contained \num{15976} records and \num{149} features. The total percentage of missing values was \SI{5.41}{\percent}, and the total percentage of NA values was \SI{20.33}{\percent}. 

To pre-process the dataset, missing values were replaced with 0, and records with more than \SI{30}{\percent} of missing values were removed (\SI{10}{\percent} of the data). For cells that contained NA values, each relevant feature was paired with another feature called \textsc{FeatureName NA} that received the value of 1 if the original feature was not relevant to the record. For example, the feature Not in Education, Employment or Training (\textsc{neet}) is not applicable to those under 16 years, resulting in the presence of NA values. Supplementary Table~S11 contains the statistics of the features before one hot encoding. 

After pre-processing, the dataset contained \num{14360} records and \num{149} features. The number of features with less than \SI{5}{\percent} of missing values is 64 while the number of features with less than \SI{20}{\percent} of NA cells is 91. After applying the one-hot encoding, the number of features in the pre-processed dataset was 363. The feature \textsc{locality decision} represents the target variable with three categories: \textsc{some action} (\SI{56.59}{\percent}), \textsc{eh support} (\SI{33.10}{\percent}), and \textsc{no action} (\SI{10.31}{\percent}). Those who received \gls{EHS} belong to the \textsc{eh support} category. Any other type of service provided by LCC to a child or signposting to an external organisation is considered as \textsc{some action}, and young people who did not receive any action are labeled in the \textsc{no action} category. The remaining input features represent educational indicators and are grouped into topics such as Absence, Exclusion, School Transfer, Free School Meal (\textsc{fsm}), Special Educational Needs and Disabilities (\textsc{send}), Pupil Referral Unit (\textsc{pru}), Home Education, Missing, Not in Education, Employment or Training (\textsc{neet}), Early Years Funding (\textsc{eyf}) and the Income Deprivation Affecting Children Index (\textsc{idaci}). A description of these features can be found in Supplementary Table~S5.

\subsection{Machine learning model evaluations} \label{mlmodels}
The aim is to identify the best performing \gls{ML} model for predicting three \textsc{locality decision} outcomes: \textsc{eh support}, \textsc{some action} and \textsc{no action}. The dataset was divided into two sets: a Training/validation set with 10\,052 records (\SI{70}{\percent}), and a Test set with 4\,308 records (\SI{30}{\percent}). Then the following \gls{ML} techniques were evaluated using stratified 10-fold \gls{CV}: Ridge Classification, Logistic Regression, Support Vector Classification (Linear and Kernel), K-Nearest Neighbors (KNN) Classifier, Gaussian Naive Bayes, Decision Tree, Random Forest Classifier, Gradient Boosting Classifier, Extreme Gradient Boosting, Ensemble Methods (AdaBoost, Catboost) and Discriminant Analysis (Linear and Quadratic). Supplementary Tables~S1--S3 show the results of evaluating the above-mentioned models for each \textsc{locality decision} outcome. 

The hyperparameter settings of each model are shown in Supplementary Table~S4 for reproducibility purposes. The best models were chosen based on their \gls{AUC}, recall, and precision scores on the validation sets. To ensure that the best models did not suffer from low variance, the performance of each best model was evaluated using the 10-fold cross-validation approach which was repeated 30 times on the train set. A random seed generator was applied to create a different sequence of values each time the k-fold cross-validation was run to ensure randomness. The test set remained the same across the 30 iterations. The average and standard deviation values across the iterations were recorded for the validation and test set.

Multi-class models were implemented but these did not perform to a satisfactory standard (see Supplementary Tables~S6-S7). As a solution, separate models were implemented for predicting each outcome (i.e.\ one model per module), and those achieved better results. Hence this paper presents the analysis of the separate models and the results of the multi-class models are found in the supplementary materials. 

\subsection{Evaluation metrics} 
The \gls{AUC}, recall and precision evaluation metrics were utilised to evaluate and compare the predictive performance of the \gls{ML} models. 
Recall (also known as sensitivity or true positive rate) measures the proportion of actual positive cases that a model correctly identifies. Precision (also known as positive predictive value) measures the proportion of positive predictions made by the model that are accurate. Recall and precision are calculated using the following expressions:\\ 
Recall = ${\mbox{TP}}/{(\mbox{TP} + \mbox{FN})}$ and Precision = $\mbox{TP}/{(\mbox{TP} + \mbox{FP})}$, where
\begin{itemize}
\item \glsdesc{TP} (TP) refers to a young person who required \textsc{eh support} or \textsc{some action} or \textsc{no action} and was predicted by the model as such.
\item \glsdesc{TN} (TN) refers to a young person who did not require \textsc{eh support} or \textsc{some action} or \textsc{no action} and was predicted as such.
\item \gls{FN} refers to a young person who required \textsc{eh support} or \textsc{some action} or \textsc{no action} and was predicted as not requiring such service. 
\item \gls{FP} refers to a young person who did not require \textsc{eh support} or \textsc{some action} or \textsc{no action} and was predicted as requiring such service.
\end{itemize}
The \gls{ROC} curve illustrates the trade-off between the \gls{TPR} and the \gls{FPR}, and the \gls{AUC} represents an aggregate metric that evaluates the classification performance of a model. The closer the \gls{AUC} is to 1, the better the performance of the classifier.
The distribution of classes \textsc{eh support}, \textsc{some action}, and \textsc{no action} in the target variable \textsc{locality decision} is imbalanced. In such cases, adjusting the decision threshold can be a useful technique for improving the performance of \gls{ML} models and reducing the occurrence of \glspl{FN}, which are often the most costly errors in imbalanced classification problems. By default, the threshold is set to 0.5. However, an appropriate threshold was chosen based on the trade-off between the costs associated with \gls{FP} and \gls{FN}. This was achieved by calculating the precision-recall curve and selecting the threshold that maximizes recall. The process of threshold selection is described in Section \textit{Threshold adjustment}. 

\subsection{Bias mitigation in \gls{ML} models}
The threshold optimizer and exponentiated gradient algorithms were applied as techniques to mitigate bias in the \gls{ML} models. The bias evaluation considered the following sensitive features: \textsc{gender}, \textsc{age at locality decision}, \textsc{attendance}, and \textsc{idaci}. \Gls{FNR} was used as a metric to mitigate bias since it represents those who would benefit from the \textsc{eh support} (or \textsc{some action} or \textsc{no action}) but were not predicted as such. The two-sample Z-test for proportions statistical test was applied to evaluate whether there is a significant difference between the \glspl{FNR} of the categories for a given sensitive feature (e.g., \textsc{gender}, \textsc{age at locality decision}, \textsc{attendance}, and \textsc{idaci}). The null hypothesis is that there is no significant difference between the two proportions (i.e.\ \glspl{FNR}), while the alternative hypothesis is that there is a significant difference between them. Under the null hypothesis, the test statistic follows a standard normal distribution and the p-value can be calculated using this distribution. If the p-value is less than the chosen significance level ($\alpha = 0.05$), the null hypothesis is rejected and it can be concluded that there is a significant difference between the two \gls{FNR} values, \textcolor{black}{and the \gls{ML} model may have been biased towards the sensitive feature under scrutiny}. In this case, bias mitigation algorithms are considered to reduce the bias for the sensitive feature. Otherwise, if the p-value is greater than the significance level ($\alpha = 0.05$), the null hypothesis is not rejected, and it can be concluded that there is not enough evidence to suggest a significant difference between the two \glspl{FNR}. In this case, it can also be concluded that the \gls{ML} model does not present bias for the sensitive feature under scrutiny.

\subsubsection{Threshold adjustment} 
\begin{figure}[htb]
\begin{subfigure}{.5\textwidth}
  \centering
  \includegraphics[width=01\linewidth]{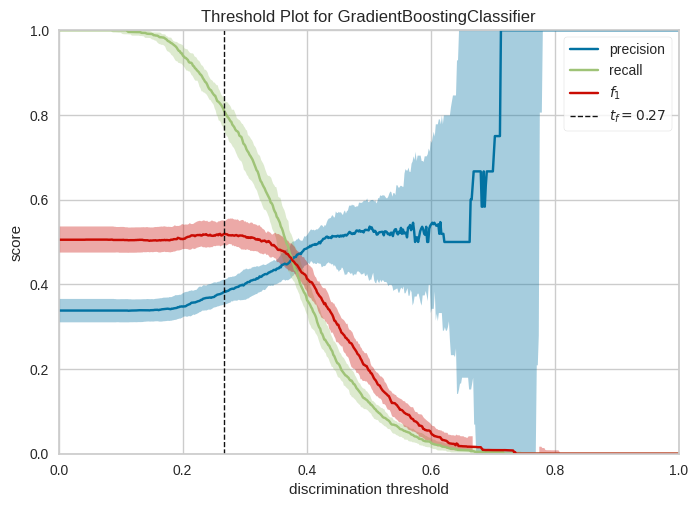}
  %\caption{Missing Values}
  \label{fig:sfig_1a}
\end{subfigure}%
\begin{subfigure}{.5\textwidth}
  \centering
  \includegraphics[width=1\linewidth]
  {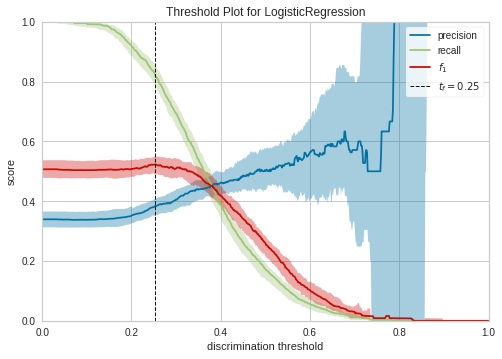}
  %\caption{Not Applicable cells}
  \label{fig:sfig_1b}
\end{subfigure}
\caption{Threshold analysis for the \gls{GBC} and \gls{LR} \textsc{eh support} models.}
\label{fig:fig_1}
\end{figure}

Threshold adjustment is applied to find the threshold that maximizes the performance of the model in terms of precision and recall. The process of adjusting the threshold for the \textsc{eh support} model is described as follows. After the model has been trained, it returns a probability score of the confidence of the model's prediction. The threshold value is then used to decide whether the prediction should be classified as \textsc{eh support} (class 1) or not (class 0). For example, without adjustment, when the model predicts a probability greater than 0.5, the record is labeled as class 1 otherwise it is labeled as class 0. With adjustment, the threshold value was adjusted based on the threshold analysis illustrated in Figure~\ref{fig:fig_1}. This analysis revealed that the optimal cutoff for classification purposes (i.e., the point with the best balance between precision and recall) occurs at a threshold of 0.27 for the \gls{GBC} model and at a threshold of 0.25 for the \gls{LR} model. The F1 curve represents the harmonic average between the precision and recall rates. The performance of these models (with and without the threshold) is then evaluated using a stratified 10-fold \gls{CV} over 30 iterations. 
Table \ref{tab2B} presents the average and standard deviation for the recall and the precision for both the \gls{GBC} and \gls{LR} models (with and without threshold). 
\begin{table}[htb]
    \centering 
    \begin{tabular}{lcc}
    \hline
        Classifier & Recall & Precision \\ \hline
        \gls{GBC} & 0.1069 $\pm$ 0.0120 & 0.5296 $\pm$ 0.0370\\ 
        \gls{GBC} (with threshold) & \textbf{0.8133 $\pm$ 0.0180} & 0.3839 $\pm$ 0.0038 \\ 
        \gls{LR} & 0.1604 $\pm$ 0.0124 & 0.5047 $\pm$ 0.0232  \\ 
        \gls{LR} (with threshold) & \textbf{0.8242 $\pm$  0.0125} & 0.3781 $\pm$ 0.0035 \\ \hline
    \end{tabular}
    \caption{\label{tab2B} Threshold adjustment for the \textsc{eh support} model: predictive performance for the \gls{GBC} and \gls{LR} models (with and without the threshold). Average and standard deviation for \gls{AUC}, recall and precision using stratified 10-fold \gls{CV} over 30 iterations.}
\end{table}
The same process followed for the \textsc{eh support} model, was followed for the other models. No threshold adjustment was needed for the \textsc{some action} model, and hence the threshold value was set to 0.5. For the \textsc{no action} model, the optimal threshold was 0.46. Supplementary Fig.~S4 illustrates these results.

\glsreset{FNR}
\section{Results}
This section describes the performance of \gls{ML} models for predicting whether a young person requires \gls{EHS}. Specifically, it evaluates binary classification models for predicting each of the following outcomes: (1) \textsc{eh support}; (2) \textsc{some action}; (3) \textsc{no action}.
The \textit{\gls{LIME}} \cite{Ribeiro2016} method is then applied to explain the model predictions by identifying those features that were most important for correct or incorrect classifications. Supplementary Table~S5 gives a description of all features used in this work. Finally, the suitability of the \textit{threshold optimiser}\cite{Hardtetal2016} and \textit{exponentiated gradient reductions}\cite{agarwaletal2018} techniques in mitigating bias are applied and these results are also reported.

\subsubsection{Can we predict whether a young person needs Early Help support? Model for \textsc{eh support}}
\begin{figure}[htb]
\centerline{\includegraphics[width=0.6\linewidth]{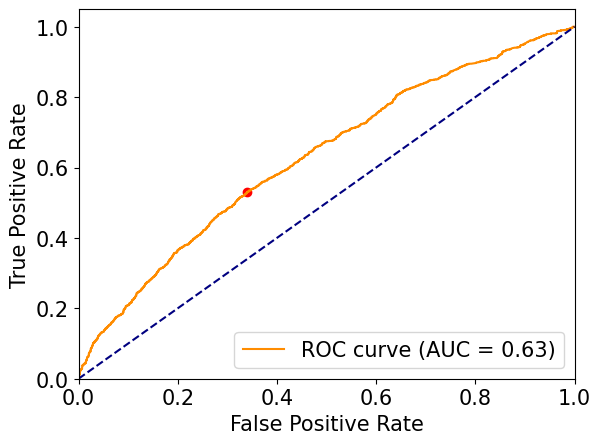}}
\caption{\textsc{eh support} Model: \gls{ROC} curve, \gls{AUC} and optimal \gls{ROC} point for the \gls{LR} classifier on the test set. 
}
\label{fig:fig_2}
\end{figure}

\textbf{Model performance.} Supplementary Table~S8 lists the validation performance of the \textit{\gls{GBC}} and \textit{\gls{LR}} models across 30 iterations. The \gls{LR} model outperformed the \gls{GBC} on the validation sets (i.e.\ across 30 iterations) with an average \gls{AUC} of 0.62 (standard deviation $\hat{\sigma} = 0.01$), an average recall of 0.82 ($\hat{\sigma} = 0.01$) and an average precision of 0.38 ($\hat{\sigma} = 0.00$). On the test set, \gls{LR} reached an \gls{AUC} of 0.63, recall of 0.83 and precision of 0.38. Figure~\ref{fig:fig_2} illustrates the \textit{\gls{ROC}} curve for the \gls{LR} classifier and Supplementary Table~S10 presents the test performance metrics and the optimal \gls{ROC} point. 

\textbf{Interpretation of results.} 
The \gls{LR} model demonstrates a moderate ability to differentiate between young people who require early help and those who do not, as indicated by the \gls{AUC} of 0.63. However, it has a relatively high recall of 0.83, meaning that it captures a good proportion of the young people who require early help. On the other hand, the precision of 0.38 suggests that the model generates a considerable number of \glspl{FP}, incorrectly identifying some young people as needing early help.

\begin{figure}[htb]
\begin{subfigure}{.5\textwidth}
  \centering
  \includegraphics[width=\linewidth]{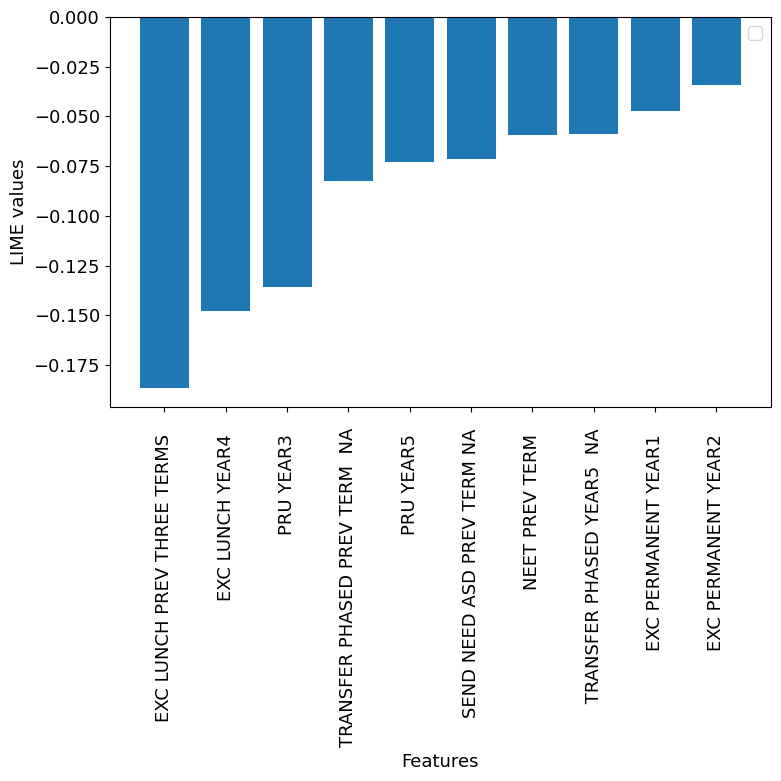}
  \caption{Most important features of the \gls{TN} group.}
  \label{fig:sfig_3a}
\end{subfigure}%
\begin{subfigure}{.5\textwidth}
  \centering
  \includegraphics[width=\linewidth]{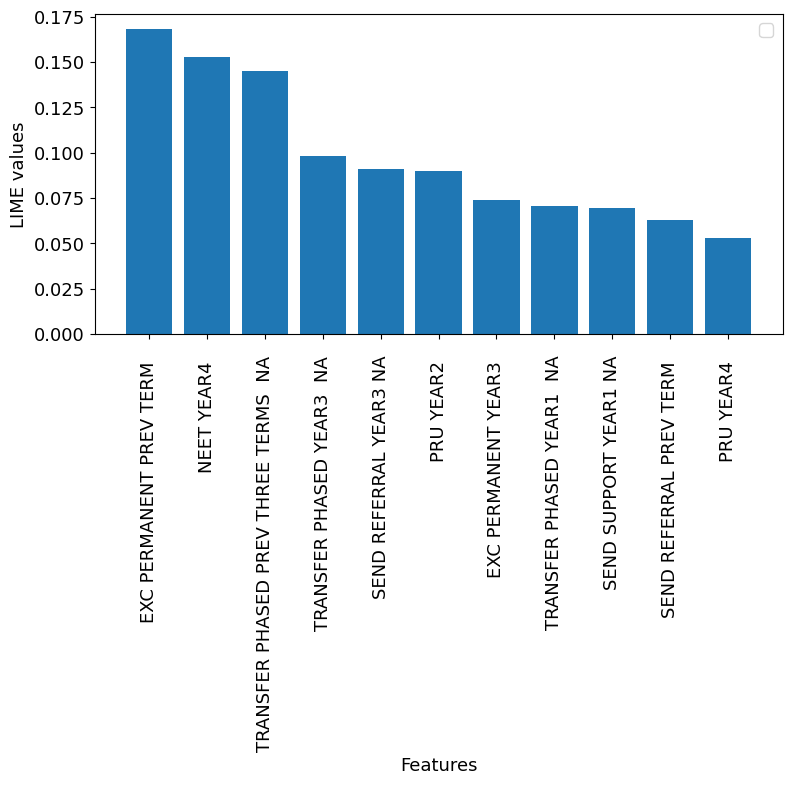}
  \caption{Most important features of the \gls{TP} group.}
  \label{fig:sfig_3b}
\end{subfigure}
\caption{\gls{LIME} analysis of young people correctly classified by the \textsc{eh support} model (\gls{LR} classifier) on the test set.}
\label{fig:fig_3}
\end{figure}

\textbf{Factor analysis.} The findings from the \gls{LIME} analysis revealed the factors related to young people correctly classified by the \textsc{eh support} model. Those who did not require \textsc{eh support}, i.e.\ the \gls{TN} group, had a median age of 14 years, compared to a median age of 8 years for those who did require \textsc{eh support}, i.e.\ the \gls{TP} group. A higher proportion of young people (\SI{91}{\percent}) who required \textsc{eh support} attended a \textit{\gls{PRU}} and received \textit{\gls{SENDEHC}} support compared to those who did not require \textsc{eh support} (\SI{77}{\percent}). Those who were not-applicable (NA) for the \textit{\gls{SEND}} or \textit{\gls{REFERRAL}} services were less likely to require \textsc{eh support}. Supplementary Fig.~S1 illustrates these findings. For those belonging to the \gls{TP} group, i.e.\ those that required \textsc{eh support} and were correctly classified by the model, the most relevant features were \textsc{permanent exclusion}, \textit{\gls{NEET}}, \textit{\gls{TRANSFER}}, \textsc{send referral} and \textsc{pru}. Figure~\ref{fig:fig_3} shows all important features for those young people that were correctly classified by the \textsc{eh support} model. The negative \gls{LIME} values in Figure~\ref{fig:sfig_3a} and the positive \gls{LIME} values in Figure~\ref{fig:sfig_3b} show the most important features of the \gls{TN} and \gls{TP} groups, respectively.

\gls{LIME} analysis also revealed that the model prioritized features with lower percentages of missing data. For example, feature \textsc{neet} (\textsc{year 4}) has 0.1\% missing values, whereas \textsc{neet} (\textsc{prev 3terms}) and \textsc{neet} (\textsc{year 1}) have \SI{3.9}{\percent} and \SI{2.3}{\percent} missing values, respectively. Another example is \textsc{pru} (\textsc{year 5}) with \SI{1.6}{\percent} missing values and \textsc{pru} (\textsc{year 1}) with \SI{6.3}{\percent} missing values (see Supplementary Table~S11). The \gls{LR} model for \textsc{eh support} had a \gls{FNR} of: 0.19 for \textsc{females}; 0.16 for \textsc{males}; and 0.33 for the category \textsc{other} which only comprises \SI{0.4}{\percent} of the data. The feature Income Deprivation Affecting Children Index (ICADI), was categorized into five classes (\textsc{idaci class}) according to the ranges (IDACI 1: [0.0, 0.2), IDACI 2: [0.2, 0.4), IDACI 3: [0.4, 0.6), IDACI 4: [0.6, 0.8), IDACI 5: [0.8, 1.0]) and the \gls{FNR} values oscillated between 0.13 and 0.22.

\begin{figure}[htb]
\begin{subfigure}{.5\textwidth}
  \centering
  \includegraphics[width=\linewidth]{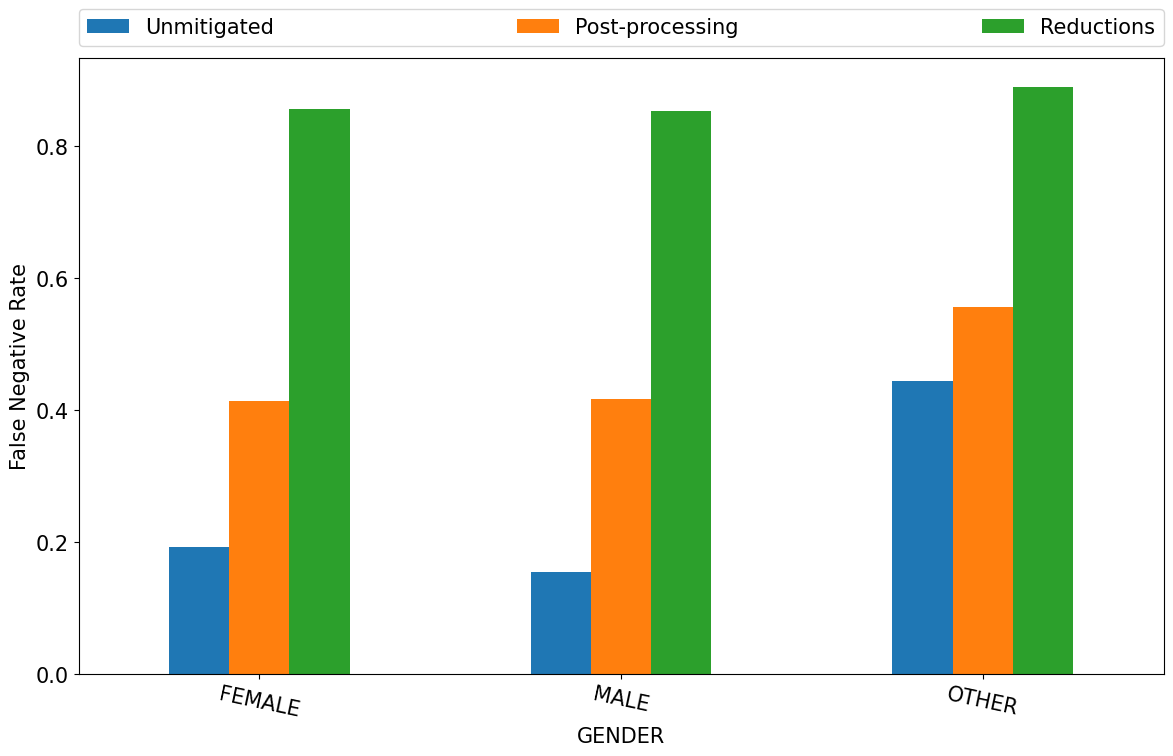}
  \label{fig:sfig_4a}
\end{subfigure}%
\begin{subfigure}{.5\textwidth}
  \centering
  \includegraphics[width=\linewidth]{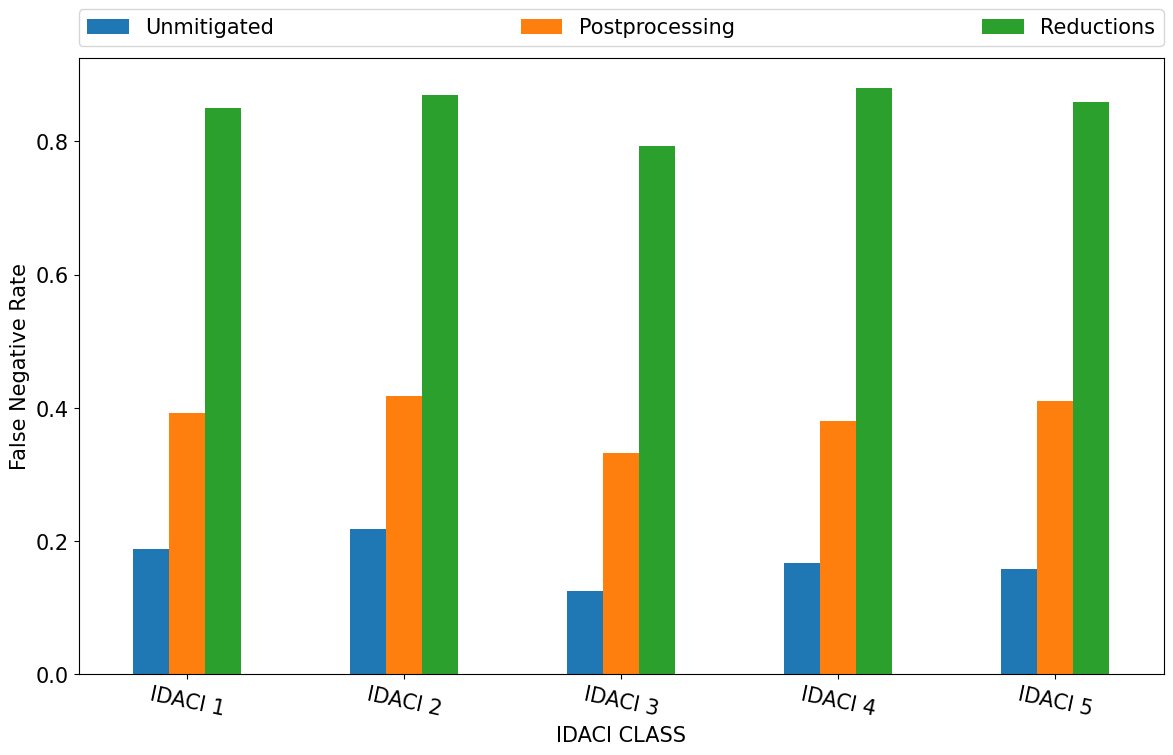}
  \label{fig:sfig_4b}
\end{subfigure}
\caption{\gls{LR} model for \textsc{eh support}. \gls{FNR} of the unmitigated and mitigated models (on the test set) for features \textsc{gender} and \textsc{idaci}. Note that `Reductions' refers to the exponentiated gradient reductions algorithm.}
\label{fig:fig_4}
\end{figure}

\textbf{Bias analysis.} Figure~\ref{fig:fig_4} shows the application of the bias-mitigation algorithms. The threshold optimizer (labelled `Post-processing') and the exponentiated gradient reductions algorithm (labelled `Reductions') reduced the \glspl{FNR} difference across categories but at the cost of an overall increase in the \gls{FNR}. 
A two-sample Z-test for proportions compared the differences of the \gls{FNR} values within the categories of the sensitive features \textsc{gender} and \textsc{idaci}. The test was not performed for the category \textsc{other} because it comprises only \SI{0.4}{\percent} of the data. According to the Z-test, there were no significant differences between the \gls{FNR} values of \textsc{female} (0.19) and \textsc{male} (0.16) as well as between most of the categories of \textsc{idaci}, where the \gls{FNR} oscillated between 0.13 and 0.22. Supplementary Table~S12 details these results and concludes that the \gls{LR} model does not present bias in the sensitive features \textsc{gender} and \textsc{idaci}.

\subsubsection{Can we predict whether Some Action is needed? Model for \textsc{some action}}
%Supplementary Table~S2 provides the predictive performance of the \gls{ML} algorithms and the \gls{GBC} reached the best \gls{AUC} and recall.

\textbf{Model performance.} The \gls{GBC} reached an average \gls{AUC} of 0.60 ($\hat{\sigma} = 0.01$), an average recall of 0.79 ($\hat{\sigma} = 0.02$) and an average precision of 0.61 ($\hat{\sigma} = 0.01$) across 30 iterations. On the test set, the model had an \gls{AUC} of 0.60,  recall of 0.81 and a precision of 0.61.

\begin{figure}[htb]
\centerline{\includegraphics[width=0.6\linewidth]{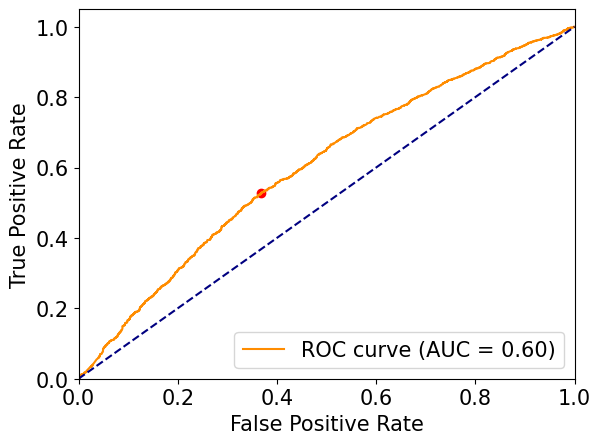}}
\caption{\textsc{some action} Model: \gls{ROC} curve, \gls{AUC} and optimal \gls{ROC} point for the \gls{GBC} on the test set.
}
\label{fig:fig_5}
\end{figure}

\textbf{Interpretation of results.} The GBC model demonstrates a modest ability to differentiate between young people who require some action and those who do not, as indicated by the \gls{AUC} of 0.60. It has a relatively high recall of 0.79, indicating its capability to capture a good proportion of young people who require action. However, the precision of 0.31 suggests that the model generates a considerable number of \glspl{FP}, incorrectly identifying a significant portion of young people as requiring action when they do not actually need it.

\begin{figure}[htb]
\begin{subfigure}{.5\textwidth}
  \centering
  \includegraphics[width=\linewidth]{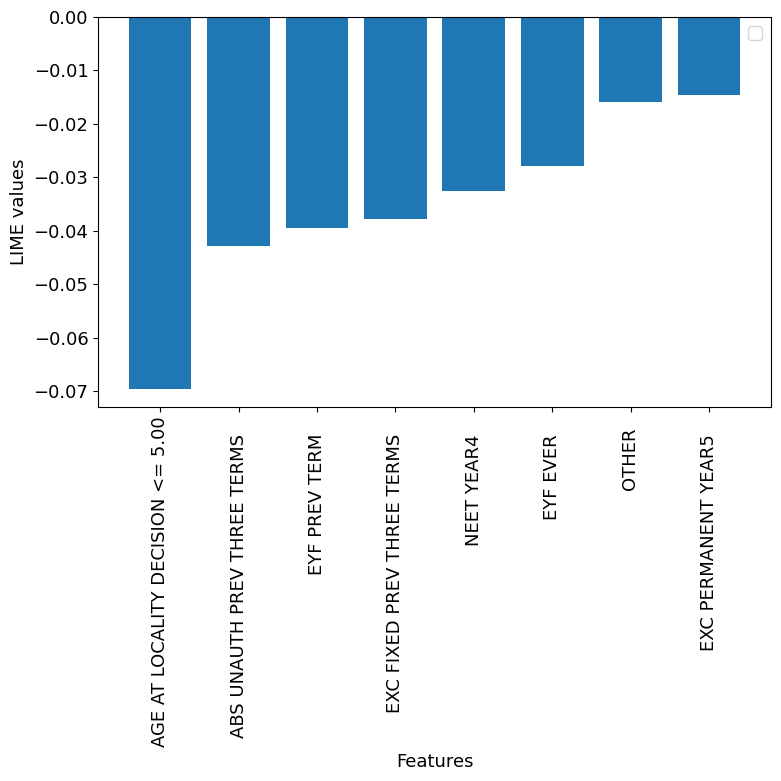}
  \caption{Most important features of the \gls{TN} group.}
  \label{fig:sfig_6a}
\end{subfigure}
\begin{subfigure}{.5\textwidth}
  \centering
  \includegraphics[width=\linewidth]{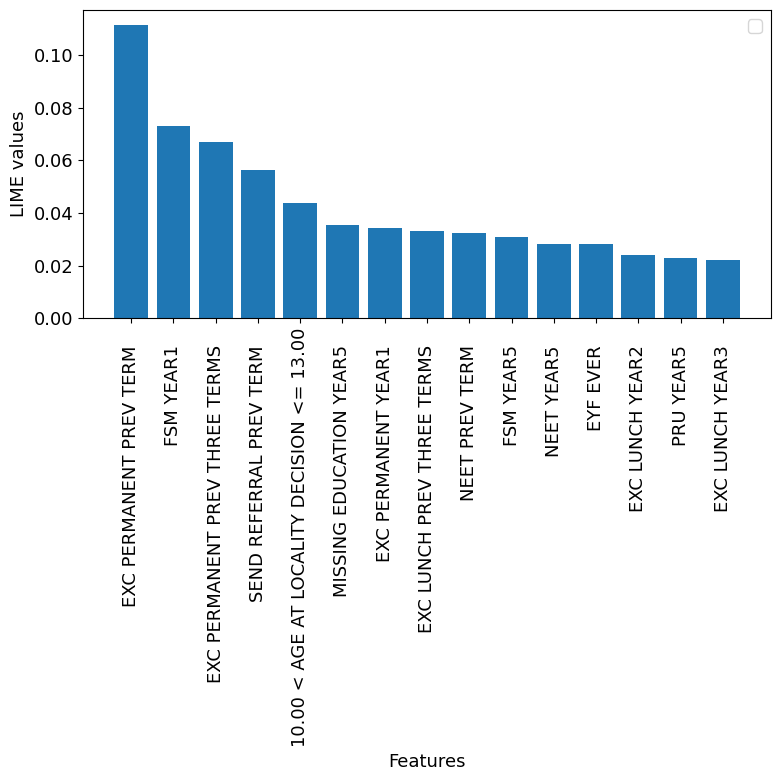}
  \caption{Most important features of the \gls{TP} group.}
  \label{fig:sfig_6b}
\end{subfigure}
\caption{\gls{LIME} analysis for those correctly classified (test set) by the \textsc{some action} model (\gls{GBC}). 
}
\label{fig:fig_6}
\end{figure}

\textbf{Factor analysis.} The \gls{LIME} analysis revealed that the median age of those who did not receive \textsc{some action} was 6 years versus 12 years for those that did. In the former group, a higher proportion received \textit{\gls{EYF}} and had \textit{\gls{FSM}}. The average number of \textsc{fixed-term exclusion} sessions was higher (0.184 average) among those who received \textsc{some action}, compared to those who did not (0.065 average). Supplementary Fig.~S2 illustrates these results. Moreover, amongst those that received \textsc{some action} and were correctly classified by the model (\gls{TP} group), the most relevant features were \textsc{permanent exclusions}, \textsc{fsm}, \textsc{send referral}, \textsc{age at locality decision} (11--13 years) and \textsc{missing education}. The negative \gls{LIME} values in Figure~\ref{fig:sfig_6a} represent the most important features of the \gls{TN} group and the positive \gls{LIME} values in Figure~\ref{fig:sfig_6b} show the most important features of the \gls{TP} group.

Young people who received \textsc{some action} had a higher average \textsc{age at locality decision} (9.3 years) compared to those who did not receive \textsc{some action} (8.5 years). The \textsc{attendance} (year 4) feature (i.e. percentage of total school attendance sessions in the academic year four years previous), also presented a difference between those who received \textsc{some action} (\SI{61.5}{\percent}) and those who did not receive \textsc{some action} (\SI{56}{\percent}). For the other features, a small difference between the groups was observed (less than \SI{2}{\percent}). Since \textsc{age at locality decision} and \textsc{attendance} (year 4) are numerical features, it was necessary to create new categorical features (\textsc{class age} and \textsc{attendance bin}, respectively) for evaluating the presence of bias. Supplementary Fig.~S5 represents histograms for both features. Feature \textsc{class age} was categorised into three groups: below 7.5 years (group A), 7.5--12.5 years (group B) and above 12.5 years (group C). Additionally, the new binary feature \textsc{attendance bin} considers two categories: $\leq 0.5$ and $> 0.5$.

The \gls{GBC} model for \textsc{some action} had an \gls{FNR} of 0.22 for the category \textsc{female} and 0.20 for the category \textsc{male}. Category \textsc{other} had the lowest \gls{FNR} of 0.08. The \gls{FNR} values between the categories of \textsc{idaci} oscillated between 0.18 and 0.25. Two-sample Z-tests were carried out to compare differences between categories in order to identify potential biases within the \gls{GBC} model. According to the Z-test, there was no significant difference between the \gls{FNR} values of \textsc{male} and \textsc{female} and between the \gls{FNR} values for most of the categories of \textsc{idaci}.
With regards to the \textsc{class age} feature, the \gls{GBC} model had an \gls{FNR} of 0.03 in group age C (above 12.5 years), 0.21 in group age B (7.5--12.5 years), and 0.38 in group age A (below 7.5 years). There was a significant difference between the \gls{FNR} values for all the groups. Similarly, for the \textsc{attendance bin} feature, the \gls{FNR} was 0.12 in the group $> 0.5$ and 0.36 in the group $\leq 0.5$. The test concluded that there was a significant difference between the \gls{FNR} values between the groups. Supplementary Table~S11 details these findings. Hence, these results reveal the presence of bias in the features \textsc{class age} and \textsc{attendance bin}.

\begin{figure}[htb]
\centerline{\includegraphics[width=0.6\linewidth]{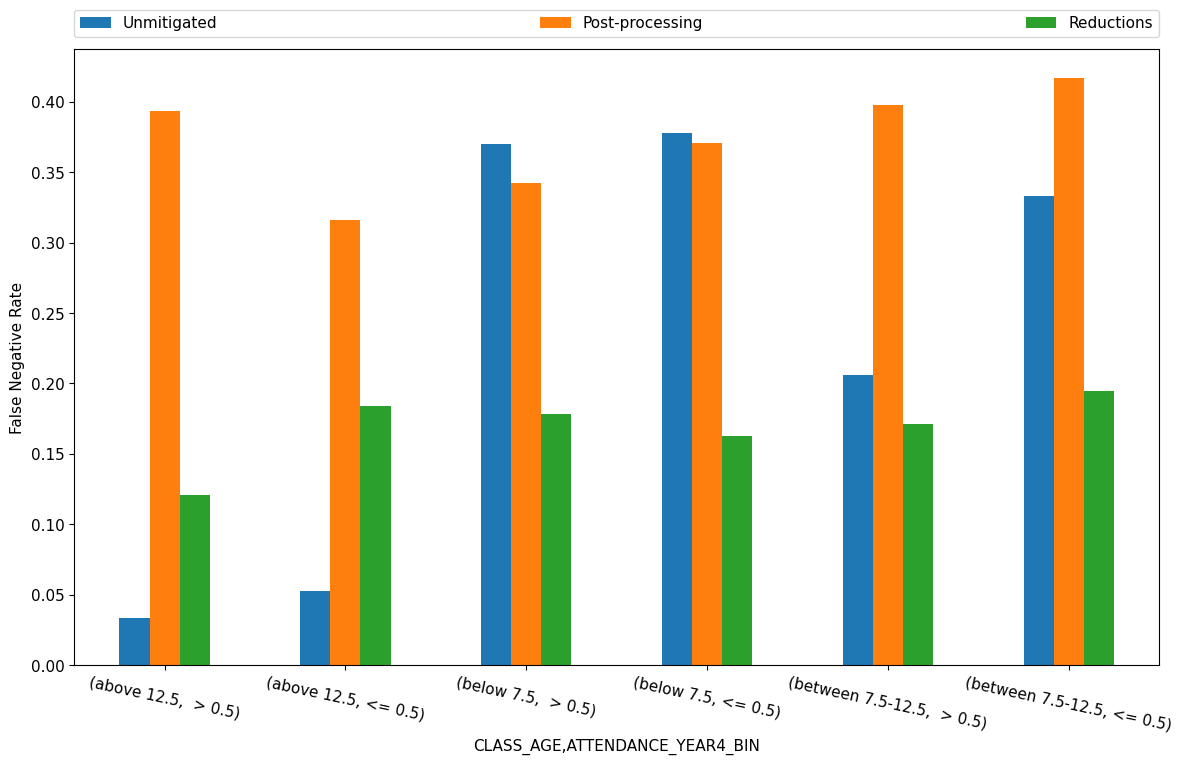}}
\caption{\gls{GBC} model for \textsc{some action}. \gls{FNR} of unmitigated and mitigated models (on the test set) for features \textsc{class age} and \textsc{attendance bin}. On the x-axis, each bracket contains two values. The first value refers to \textsc{class age} and the second refers to \textsc{attendance bin}. 
}
\label{fig:fig_7}
\end{figure}

\textbf{Bias analysis.} The use of the bias mitigation post-processing algorithm threshold optimizer reduced the gap between the categories, but it resulted in an increase in the \gls{FNR} by more than 0.30 in all groups. On the other hand, the exponentiated gradient reductions algorithm decreased the \gls{FNR} in most groups and produced the closest \gls{FNR} between the groups, giving a better outcome. Figure~\ref{fig:fig_7} illustrates these findings. The \gls{GBC} with exponentiated gradient reductions algorithm had a precision of 0.59 and a recall of 0.85 on the test set, wheread the \gls{GBC} unmitigated model had a precision of 0.61 and recall of 0.80. From these results, it can observed that the \gls{GBC} model with the exponentiated gradient reductions algorithm outperforms the \gls{GBC} unmitigated model in terms of recall (0.85 > 0.80). However, the \gls{GBC} unmitigated model has a slightly higher precision (0.61 > 0.59). Moreover, the use of the exponentiated gradient algorithm reduced the overall \gls{FNR} from 0.21 to 0.15.

\subsubsection{Can we predict whether \textsc{no action} is needed? Model for \textsc{no action}}
\begin{figure}[htb]
\centerline{\includegraphics[width=0.6\linewidth]{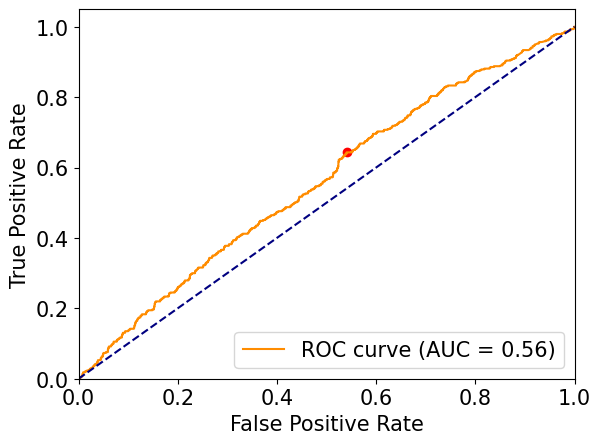}}
\caption{\textsc{no action} Model: \gls{ROC} curve, \gls{AUC} and optimal \gls{ROC} point for the \gls{LR} classifier on the test set.}
\label{fig:fig_8}
\end{figure}

\textbf{Model performance.} Supplementary Table~S9 provides the validation performance of the \gls{ML} models over 30 iterations. The validation results of the LR model reached an average AUC of 0.56 ($\hat{\sigma}$ = 0.01), an average recall of 0.60 ($\hat{\sigma}$ = 0.03), and an average precision of 0.11 ($\hat{\sigma}$ = 0.01). The presence of imbalanced classes reflects the lower values for recall and precision. On the test set the model had an \gls{AUC} of 0.56, recall of 0.63 and a precision of 0.12. None of the models performed well on this task. The presence of imbalanced classes reflects their low performance. Figure~\ref{fig:fig_8} illustrates the \gls{ROC} curve and Supplementary Table~S10 presents the predictive performance metrics and the optimal \gls{ROC} for the \gls{LR} classifier on the test set.

\textbf{Interpretation of results.}  The \gls{LR} model has a relatively low \gls{AUC} score, suggesting poor performance in distinguishing between positive and negative instances. Although it exhibits a decent recall rate, capturing 63\% of the young people who require no action, its precision of 12\% is quite low, indicating a high number of \glspl{FP}. These results suggest that the model may have difficulty accurately identifying individuals who require no action.

\begin{figure}[htb]
\begin{subfigure}{.5\textwidth}
  \centering
  \includegraphics[width=\linewidth]{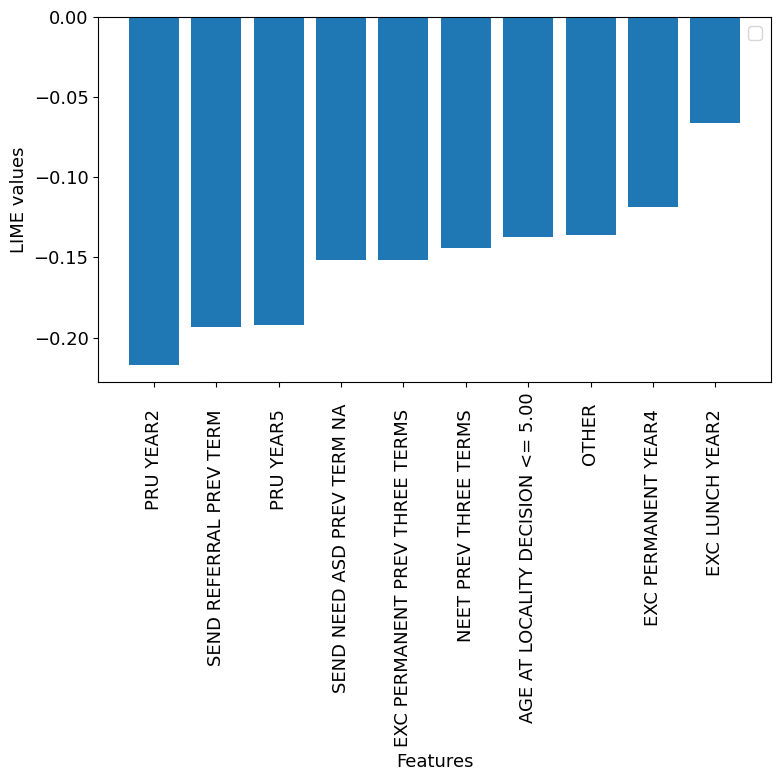}
  \caption{Important features of the \gls{TN} group.}
  \label{fig:sfig_9a}
\end{subfigure}%
\begin{subfigure}{.5\textwidth}
  \centering
  \includegraphics[width=\linewidth]{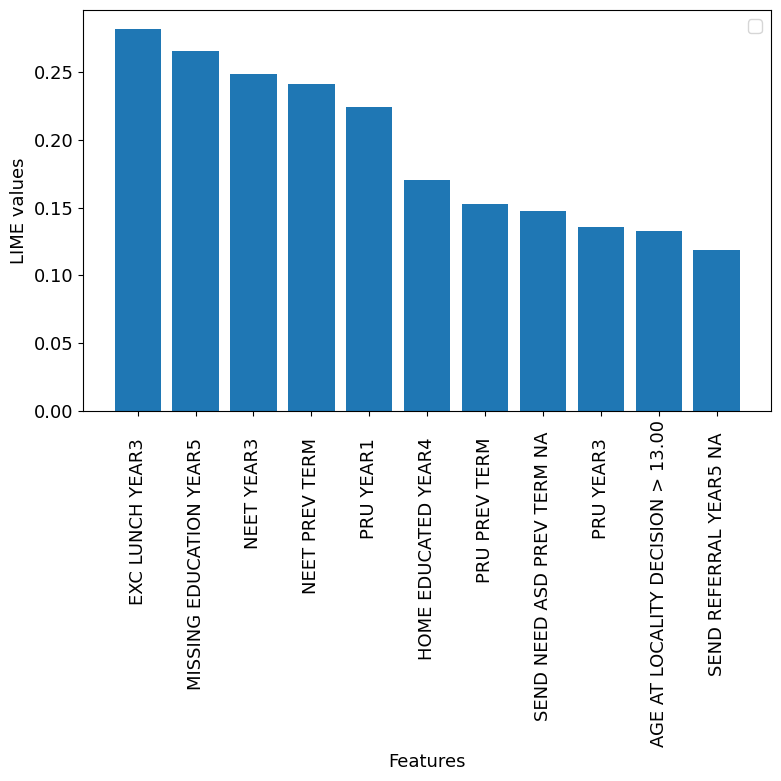}
  \caption{Important features of the \gls{TP} group.}
  \label{fig:sfig_9b}
\end{subfigure}
\caption{LIME analysis for young people correctly classified by the \textsc{no action} model (LR classifier) on the test set. 
}
\label{fig:fig_9}
\end{figure}

\textbf{Factor analysis.} The \gls{LIME} results for the \gls{LR} classifier are illustrated in Figure~\ref{fig:fig_9}. The negative \gls{LIME} values in Figure~\ref{fig:sfig_9a} represent the most important features of the \gls{TN} group and the positive \gls{LIME} values in Figure~\ref{fig:sfig_9b} shows the most important features of the \gls{TP} group. For those in the \textsc{no action} category that were correctly classified by the model (\gls{TP} group), the most relevant features were \textsc{exclusion lunchtime}, \textsc{missing education}, \textsc{neet}, \textsc{pru} and \textsc{home educated}.

Moreover, it was identified that a high proportion of those who had NA values in the \textsc{send need} feature did not belong to the \textsc{no action} category (hence they had received \textsc{eh support} or \textsc{some action}). This result aligns with the finding obtained by the \textsc{eh support} model. Young people in the \gls{TN} group had a median age of 9 years, compared to a median of 12 years in the \gls{TP} group. The box plot suggests no difference between the age of those belonging to the \gls{TN} and \gls{TP} groups (see Supplementary Fig.~S3). The \gls{LR} model for \textsc{no action} had an \gls{FNR} of 0.36 for the category \textsc{female} and 0.39 for the category \textsc{male}. The \gls{FNR} values between the categories of \textsc{idaci} oscillated between 0.30 and 0.41. According to the two-sample Z-test, there was no significant difference between the \gls{FNR} values of \textsc{female} and \textsc{male} and between all the categories of \textsc{idaci}. This suggests that the \gls{LR} model is not biased with respect to these features. Regarding the feature \textsc{class age}, the \gls{LR} model had an \gls{FNR} of 0.36 in group age A (below 7.5 years), 0.56 in group age B (7.5--12.5 years) and 0.23 in age group C (above 12.5 years). The two-sample Z-test concluded that there was a significant difference between the \gls{FNR} values between all groups suggesting the presence of bias in the feature \textsc{class age}. Supplementary Table~S12 details these results.

\begin{figure}[htb]
\centerline{\includegraphics[width=0.6\linewidth]{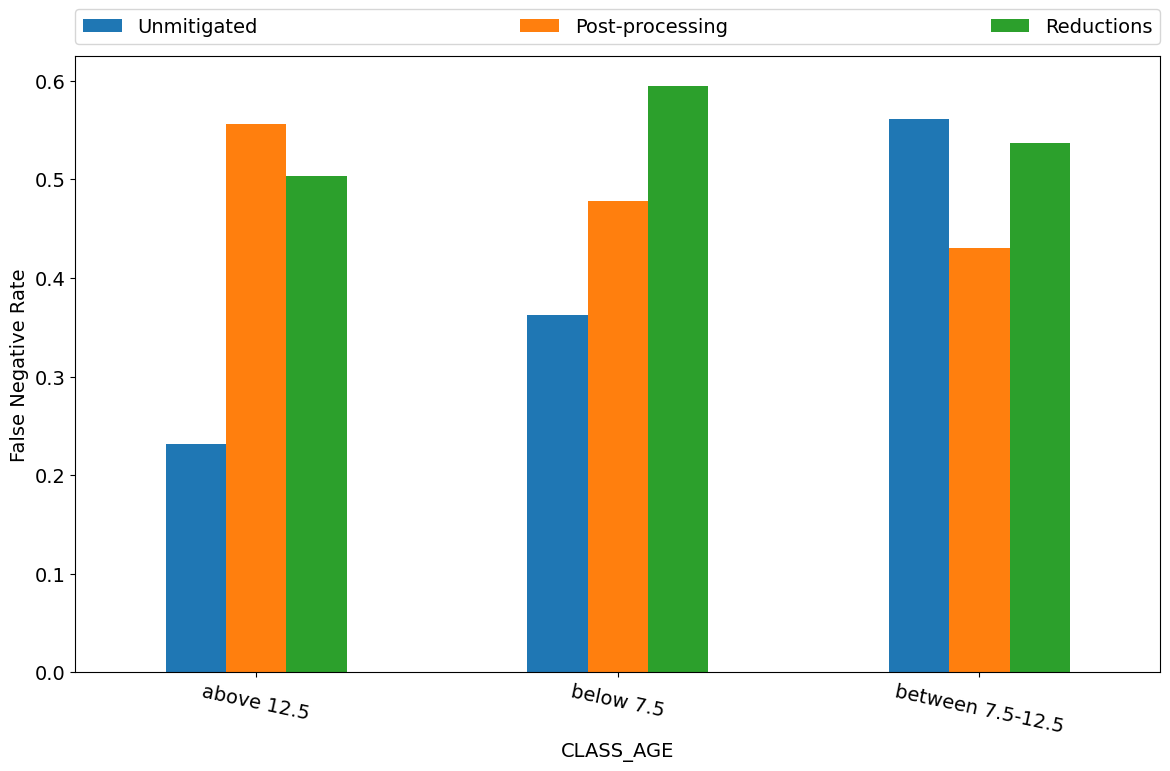}}
\caption{\gls{LR} model for \textsc{no action}. \gls{FNR} of unmitigated and mitigated models (on the test set) for \textsc{class age}.}
\label{fig:fig_10}
\end{figure}

The use of the bias mitigation algorithms, threshold optimizer (post-processing) and exponentiated gradient (reductions) decreased the difference between the categories but resulted in an increase in the \gls{FNR} for all groups, which is not ideal for classification purposes. Figure~\ref{fig:fig_10} illustrates these findings.
The \gls{LR} classifier demonstrated the best predictive performance on the test set and correctly identified \SI{61}{\percent} of all young people within the \textsc{no action} category. However, the bias mitigation algorithms did not reduce the \gls{FNR} and the presence of a strong imbalance in the target variable affected the predictive performance of the model.

\section{Discussion}
\textbf{Overview.} For the social care task described in this paper, models were developed to determine whether machine learning models can assist human decision-makers with regard to identifying families whose young people may require \gls{EHS}. Young people who could benefit from \gls{EHS} but are not identified or offered such services can be disadvantaged by the social care system. Therefore, it was important to identify features of those that could be disproportionately negatively impacted by the \gls{ML} models as a strategy for understanding and communicating the limitations of \gls{ML} models. Since the dataset was sparse and noisy, adequate data treatment was required prior to the use of the \gls{ML} models. Imputation techniques were considered as well as the use of one-hot encoded features, allowing the \gls{ML} algorithms to distinguish between NA, missing, and filled-out cells. The pre-processed dataset was thereafter utilised for data analysis and \gls{ML} tasks. 

\textbf{Model testing.} During testing, while the models show some capability in capturing young people who require intervention or early help, they also generate a significant number of \glspl{FP}, incorrectly identifying individuals who do not actually need early help referral. This indicates room for improvement in terms of precision and overall performance in accurately identifying those in need of action or help.

\textbf{Bias analysis.} The bias analysis revealed that sensitive features \textsc{gender} and \textsc{idaci} did not bias the model with regards to predicting \textsc{locality decision} (i.e.\ \textsc{eh support}, \textsc{some action}, \textsc{no action}). However, the analysis identified \textsc{age at locality decision} and \textsc{attendance} (year 4) as sensitive features and the difference between the \gls{FNR} values for some groups was statistically significant. For example, for young people of age below 7.5 years ($\mathrm{FNR} = 0.38$) and for those of age 7.5--12.5 years ($\mathrm{FNR} = 0.03$). The use of bias mitigation algorithms reduced the \gls{FNR} in these groups and improved the predictive performance of the models \textsc{eh support} and \textsc{some action}. The data imbalance in the \textsc{no action} category affected the model's predictive performance. For those that were correctly classified by the \textsc{eh support} model (and hence who did not require \textsc{eh support}) had a median age of 14 years, compared to a median age of eight years for those who did require \textsc{eh support}. A higher proportion of young people who required \textsc{eh support} attended \textsc{pru} services or received \textsc{send} support. However, the median age for those who did not receive \textsc{some action} was 6 years and young people with a median age of 12 years received \textsc{some action}. Furthermore, in the group of those that did not receive \textsc{some action}, it was identified that a higher proportion received benefits such as \textsc{eyf} or \textsc{fsm}. Although a variety of \gls{ML} algorithms and bias mitigation techniques were considered, fairness is a socio-technical challenge, and therefore mitigations are not all technical and need to be supported by processes and practices. The use of sensitive features including demographic information during the analysis of the \gls{ML} results can enhance the understanding of model behaviour and aid with the identification of groups that could be subject to bias. It is important to assess \gls{ML} performance differences across groups and the likelihood of bias.

\textbf{Conclusion.} The findings from our study demonstrate that \gls{ML} has the potential to support decision-making in social care, and the results highlight that further research is needed in developing methods that work on such complex datasets. In particular, further research is needed in developing methods and strategies for dealing with missing, uncertain, and sparse data; and in developing \gls{ML} models that can provide clear explanations for their predictions. Research is required about how to best visualise and communicate outputs of \gls{ML} models to end-users in a way that supports decision-making.

\section{Limitations}
The limitations of the study are as follows. 
\begin{itemize}
    \item A limitation of this study is that a decision was taken early on to restrict to young people aged 18 and under at the point of assessment. However, this does leave a blind spot in the study of young people (n=\num{23420}) aged 18 and under that were never referred for assessment (but possibly should have been). 
\item The original dataset contained information about \textsc{first language}. However, this feature was excluded from the study due to a high proportion (\SI{29}{\percent}) of missing values. 
\item The dataset also contained information on \textsc{ethnicity}. Analysis of the \gls{ML} results revealed that the performance was similar across the groups and models (see Supplementary Table~S13). Ethnicity was categorised as either \textsc{white} or \textsc{non-white} due to the lower frequency of other ethnic categories (see Supplementary Table~S14). The team is currently in the process of pursuing a follow-up collaboration with \gls{LCC} to expand this study. Future work involves analysing the data of primary and secondary school young people who require \textsc{eh support} and incorporating new demographic features to uncover new insights and findings. 
\item Imputation methods were explored in our previous study and not reported in this paper. None of the imputation methods explored were suitable for imputing missing values in this dataset due to their sensitivity, and therefore it was considered ethical not to impute the missing values. Instead the missing not at random values were treated using one-hot encoding. Further study is needed to develop algorithms suitable for imputing random missing values. 
\item Class imbalance appears to be a contributor to the models' low precision values. The \textsc{some action} model that was trained on a near-balanced dataset achieved the highest precision (i.e. 61\%) compared to the other models that were not trained on balanced datasets (\textsc{eh support}: 38\% precision; and \textsc{no action}: 12\% precision). 
\item This paper focuses on the needs and characteristics of individuals and models their service requirements accordingly. This approach does not take into consideration the broader family group and how those complex interrelationships may impact on the requirements of each individual within the group. With \gls{EHS} being a whole family intervention service there is a future piece of work to understand those interrelationships and identify requirements at a family level.
\end{itemize}
\section{Impact on social care }
The findings of this paper provide an entry point for local authorities into using \gls{AI} to support the optimal provision of \gls{EHS}. Whilst acknowledging the limitations and the need to approach the implementation very carefully, this is a positive step in the long road to incorporating \gls{AI} into the decision-making process within \gls{EHS} and potentially the broader remit of Children's social care. 
At this early stage, a suitable use-case for the model would be to provide additional, data-driven support to the triage process, placing the \gls{AI} outputs alongside the descriptive referral case notes and information collected by front-line workers. With the focus throughout this paper being on providing explainable \gls{AI} models, a softer benefit would be to expand confidence and understanding of \gls{AI} within practitioners and the benefits it could bring to their daily decision-making. 

In addition to providing a more complete understanding of the needs of those referred to \gls{EHS} the model also has the potential to help identify those that may be in need of support but that have not been referred. With the focus on \gls{EHS} being to provide support and intervention before issues escalate, identifying this group and acting accordingly would be expected to reduce the requirement for higher-intensity support later on. With the current limitations of the model such an approach would need careful consideration as to how that fitted into the existing referral processes. It is not considered justifiable, certainly at this stage, for referrals and allocation of provision to be driven by \gls{AI}.

%%%%%%%%%%%%%%%%%%%%%%%%%%%%%%%%%%%%%%%%%%
\vspace{6pt} 

%%%%%%%%%%%%%%%%%%%%%%%%%%%%%%%%%%%%%%%%%%

\section*{Declarations}
\begin{enumerate}
\item \textbf{Competing interests.} The authors declare no competing interests.
\item \textbf{Data availability.} The data cannot be made publicly available because the demographic features in combination with other features may reveal a child's identify \cite{sweeney2000simple}. Statistical information about the data has been provided in the Supplementary material. The code and a sample set of records are available under the Github repository \url{https://github.com/gcosma/ThemisAIPapers-Code}
\item \textbf{Ethical approval.} This article does not contain any studies with human participants performed by any of the authors. 
\item \textbf{Informed consent.} Informed consent was obtained from all participants and/or their legal guardians.

\end{enumerate}

\section*{Acknowledgements}
The project is funded via The Higher Education Innovation Fund (HEIF) (No.\ EPG141) of Loughborough University. We acknowledge the expert and financial support of Leicestershire County Council.

\section*{Author contributions statement}
All authors contributed to the design of the study. J.B. extracted and prepared the dataset. E.A.L.N. and G.C. conceived the experiments. E.A.L.N. conducted the experiment(s). E.A.L.N. and G.C. analysed the results. E.A.L.N, G.C., A.F. and J.B. interpreted the findings. G.C., A.F. and J.M supervised the project. E.A.L.N. drafted the initial version of the manuscript. All authors reviewed and revised the manuscript.

\section*{Funding}
The project is funded via The Higher Education Innovation Fund (HEIF) (No.\ EPG141) of Loughborough University. We acknowledge the expert and financial support of Leicestershire County Council.

%%%%%%%%%%%%%%%%%%%%%%%%%%%%%%%%%%%%%%%%%%

\newpage
%%%%%%%%%%%%%%%%%%%%%%%%%%%%%%%%%%%%%%%%%%
\section*{Appendix}
\begin{table}[h]
    \centering
    \begin{tabular}{c|c|c|c}
    \hline
        Classifier & AUC & Recall & Precision \\ \hline
        Gradient Boosting & 0.6330 & 0.1201 & 0.5103  \\ %\hline
        AdaBoost & 0.6250 & 0.1541 & 0.4975  \\ %\hline
        CatBoost & 0.6240 & 0.1772 & 0.5022  \\ %\hline
        Logistic Regression & 0.6223 & 0.1722 & 0.5025  \\ %\hline
        Linear Discriminant & 0.6196 & 0.1852 & 0.4886  \\ %\hline
        Random Forest & 0.5976 & 0.1999 & 0.4291  \\ %\hline
        K Neighbors & 0.5609 & 0.2503 & 0.4278 \\ %\hline
        Naive Bayes & 0.5763 & 0.8614 & 0.3549  \\ %\hline
        Extra Trees & 0.5753 & 0.2344 & 0.4047  \\ %\hline
        Decision Tree & 0.5396 & 0.3801 & 0.3961  \\ %\hline
        Quadratic Discriminant & 0.5087 & 0.9698 & 0.3415  \\ %\hline
        Dummy & 0.5000 & 0.0000 & 0.0000  \\ %\hline
        SVM (Linear Kernel) & 0.0000 & 0.2725 & 0.4656  \\ %\hline
        Ridge & 0.0000 & 0.1655 & 0.5019  \\ \hline
    \end{tabular}
    \captionsetup{labelformat=empty}
    \caption{\label{tab1} \textbf{Table S1.} \textsc{eh support}: Validation performance of ML models. Single run stratified 10-fold CV.}   
\end{table}

\begin{table}[htb]
    \centering
    \begin{tabular}{c|c|c|c}
    \hline
        Classifier & AUC & Recall & Precision \\ \hline
        Gradient Boosting & 0.6002 & 0.7922 & 0.6043 \\ %\hline
        Logistic Regression & 0.5997 & 0.7629 & 0.6050 \\ %\hline
        CatBoost & 0.5973 & 0.7508 & 0.6077 \\ %\hline
         Linear Discriminant & 0.5960 & 0.7637 & 0.6047 \\ %\hline
        AdaBoost & 0.5938 & 0.7427 & 0.6093 \\ %\hline
        Random Forest & 0.5689 & 0.7162 & 0.5985 \\ %\hline
        Naive Bayes & 0.5639 & 0.7508 & 0.5901 \\ %\hline
        Extra Trees & 0.5497 & 0.6751 & 0.5929 \\ %\hline
        K Neighbors & 0.5311 & 0.6880 & 0.5828 \\ %\hline
        Quadratic Discriminant & 0.5229 & 0.8732 & 0.5759 \\ %\hline
        Decision Tree & 0.5217 & 0.5877 & 0.5857 \\ %\hline
        Dummy & 0.5000 & 1.0000 & 0.5630 \\ %\hline
        Ridge & 0.0000 & 0.7657 & 0.6049 \\ %\hline
        SVM (Linear Kernel) & 0.0000 & 0.4290 & 0.6083 \\ \hline
    \end{tabular} 
     \captionsetup{labelformat=empty}
     \caption{\label{tab2} \textbf{Table S2.} \textsc{some action}: Validation performance of ML models. Single run stratified 10-fold CV.}
\end{table}

\begin{table}[htb]
    \centering
    \begin{tabular}{c|c|c|c}
    \hline
        Classifier & AUC & Recall & Precision \\ \hline
        Gradient Boosting & 0.5715 & 0.0029 & 0.0833 \\ %\hline
        CatBoost & 0.5708 & 0.0057 & 0.3000 \\ %\hline
        AdaBoost & 0.5612 & 0.0029 & 0.1250 \\ %\hline
        Logistic Regression & 0.5505 & 0.0029 & 0.1500 \\ %\hline
        Linear Discriminant & 0.5490 & 0.0043 & 0.0658 \\ %\hline
        Random Forest & 0.5425 & 0.0187 & 0.1441 \\ %\hline
        Extra Trees & 0.5221 & 0.0346 & 0.1400 \\ %\hline
        K Neighbors & 0.5147 & 0.0043 & 0.0389 \\ %\hline
        Decision Tree  & 0.5043 & 0.1341 & 0.1126 \\ %\hline
        Naive Bayes & 0.5032 & 0.9135 & 0.1009 \\ %\hline
        Dummy & 0.5000 & 0.0000 & 0.0000 \\ %\hline
        Quadratic Discriminant & 0.4891 & 0.9539 & 0.0965 \\ %\hline
        SVM (Linear Kernel) & 0.0000 & 0.0290 & 0.0339 \\ %\hline
        Ridge & 0.0000 & 0.0000 & 0.0000 \\ \hline
    \end{tabular} 
     \captionsetup{labelformat=empty}
     \caption{\label{tab3} \textbf{Table S3.} \textsc{no action}: Validation performance of ML models. Single run stratified 10-fold CV.}
\end{table}

\begin{landscape}
\begin{table}[htb]
    \centering
    \begin{tabular}{c|c}
    \hline
        \textbf{Classifier} & \textbf{Hyperparameters} \\ \hline
        Gradient Boosting & Learning rate = 0.1; Max depth = 3; Estimators = 100 \\ \hline
        CatBoost & Learning rate = 0.02; L2 regularization = 3; Depth = 6; Random Strength = 1\\ \hline
        AdaBoost & Learning rate = 1.0; Estimators = 50\\ \hline
        Logistic Regression & C = 1.0; Penalty = (l1, l2); Solver = (lbfgs, liblinear)  \\ \hline
        Linear Discriminant & Components = None; Solver = svd  \\ \hline
        Random Forest & Criterion = gini; Max depth = None; Estimators = 100\\ \hline
        Extra Trees & Criterion = gini; Max depth = None; Estimators = 100\\ \hline
        Decision Tree  & Criterion = gini; Max depth = None; Splitter = best \\ \hline
        K Neighbors & Neighbors = 5, Metric = minkowski,  Weights = uniform \\ \hline
        Naive Bayes & Priors = None\\ \hline
        Quadratic Discriminant & Priors = None  \\ \hline
        Dummy & Strategy = prior \\ \hline
        SVM  & Kernel = Linear; C = 0.15 \\ \hline
        Ridge & Alpha = 1.0; Solver = auto\\ \hline
    \end{tabular} 
     \captionsetup{labelformat=empty}
     \caption{\label{tabhp} \textbf{Table S4.} Hyperparameter settings of classifiers: Learning rate = determines the step size at each iteration; Estimators = number of trees; Depth = maximum depth of the tree (if None, nodes are expanded until all leaves are pure or until all leaves contain less than min\_samples\_split samples); L2 regularization = coefficient at the regularization term of the cost function; Random Strength = amount of randomness when the tree structure is selected (avoid overfitting in the model);  C = inverse of the regularisation strength; Solver = optimisation algorithm (if auto, chooses the solver automatically based on the type of data); Components = number of components for dimensionality reduction (if None, will be set to $\min(n\_classes - 1,$ $ n\_features)$); Criterion = function used to measure the quality of a split; Splitter = the strategy used to choose the split at each node; Kernel = the kernel type to be used, e.g. linear, polynomial; Neighbours = the number of nearest records to consider when making a prediction; Weights = how to weight neighbors when making a prediction; Metric = Distance calculation algorithm, Priors = prior probabilities of the classes; Strategy = strategy to use to generate predictions;  Kernel = the kernel type to be used, e.g. linear, polynomial; Alpha = constant that multiplies the L2 term, controlling regularization strength. All classifiers used a random\_state parameter equal to 123 (controls the randomness of the estimator)}.
\end{table}
\end{landscape}
\begin{landscape}
\begin{table}[htb]
    \centering
    {\footnotesize
    \begin{tabular}{p{1.5in}|c|c| p{0.35\linewidth}}
    \hline
        Feature & Label & Type & Description \\ \hline
        Locality decision & \textsc{locality decision} & Categorical & Type of early help service received by a young person (\textsc{eh support}, \textsc{some action} or \textsc{no action}). \\
        Gender & \textsc{gender} & Categorical & Gender of a person. \\ 
        Age at locality decision & \textsc{age at locality decision} & Numeric & Age when a locality decision was made. \\
        Locality decision incidence & \textsc{locality decision incidence} & Numeric & Frequency of locality decision incidence. \\ %\hline
        Income deprivation affecting children index & \textsc{idaci} & Numeric & Proportion of young people living in income-deprived families in the home postcode of the individual. \\
        Authorised absence$^{*}$ & \textsc{abs auth} & Numeric & Percentage of total school attendance sessions classified as authorised absence \\
        Unauthorised absence$^{*}$ & \textsc{abs unauth} & Numeric & Percentage of total school attendance sessions classified as unauthorised absence \\
        Attendance$^{*}$ & \textsc{attendance} & Numeric & Percentage of total school attendance sessions.\\
        Permanent exclusions$^{*}$ & \textsc{exc permanent} & Numeric & Number of permanent school exclusions.\\
        Lunchtime exclusions$^{*}$ & \textsc{exc lunch} & Numeric & Number of lunchtime school exclusion sessions (\textsc{cnt}) and lunchtime school exclusion sessions as a percentage of total school attendance sessions (\textsc{pct}).\\
        Fixed term exclusions$^{*}$ & \textsc{exc fixed} & Numeric & Number of fixed term school exclusion sessions (\textsc{cnt}) and fixed term school exclusion sessions as a percentage of total school attendance sessions (\textsc{pct}).\\
        School transfer phased$^{*}$ & \textsc{transfer phased} & Numeric & Number of phased school transfers. Phased meaning they naturally progressed from primary school to secondary school.\\
        School transfer mid-term$^{*}$ & \textsc{transfer midterm} & Numeric & Number of mid-term school transfers. Mid-term means that the transfer happened outside of the normal phased transfer pattern.\\
        Pupil referral unit$^{*}$ & \textsc{pru} & Binary & Indicates if a young person was attending a \textsc{pru} provision.\\
        Home education$^{*}$ & \textsc{home educated} & Binary & Indicates if a young person was home educated.\\
        Missing children$^{*}$ & \textsc{missing education} & Binary & Indicates if a young person was reported missing.\\
        Free school meal$^{*}$ & \textsc{fsm} & Binary & Indicates if a young person was eligible for \textsc{fsm}.\\
        Not in education, employment or training$^{*}$ & \textsc{neet} & Binary & Indicates if a 16-24-year old young person was \textsc{neet}.\\
        Early years funding$^{*}$ & \textsc{eyf} & Binary & Indicates if a \textsc{eyf} funding was claimed for a two-year-old child.  \textsc{eyf ever} indicates if an \textsc{eyf} funding was claimed for a two-year-old at any point.\\
        Special Educational Needs and Disabilities$^{*}$ & \textsc{send} & Binary & Indicates if a young person was referred for assessment (\textsc{send referral}) and, if so, was the outcome for SEN Support (\textsc{send support}) or a Education, Health and Care (EHC) Plan (\textsc{send ehc}).  Predominant need (\textsc{send need}) is included independently of the level of support provided.\\
        \hline
    \end{tabular}
    }
   \captionsetup{labelformat=empty}
     \caption{\label{tab5_v2} \textbf{Table S5.} Features description. Features marked with asterix ($*$) are computed for the previous term before the locality decision (\textsc{prev term}), the previous three terms (\textsc{prev 3 terms}) and the academic year previous (\textsc{year 1}), two years previous (\textsc{year 2}), three years previous (\textsc{year 3}), four years previous (\textsc{year 4}) and five years previous (\textsc{year 5}).} 
\end{table}
\end{landscape}
\begin{figure}[htb]
\begin{subfigure}{.5\textwidth}
  \centering
  \includegraphics[width=\linewidth]{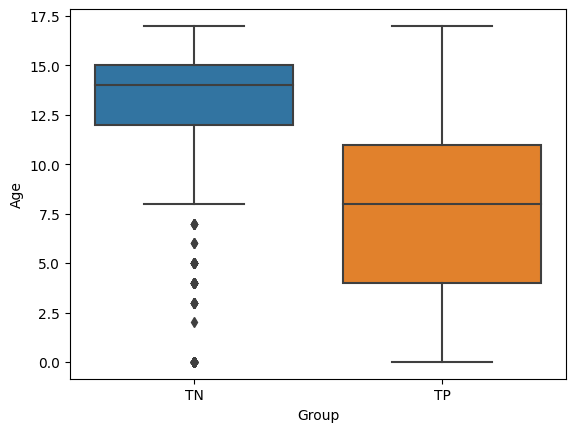}
%  \caption{AGE\_AT\_LOCALITY\_DECISION}
  \label{fig:sfig_s1_age}
\end{subfigure}%
\begin{subfigure}{.5\textwidth}
  \centering
  \includegraphics[width=\linewidth]{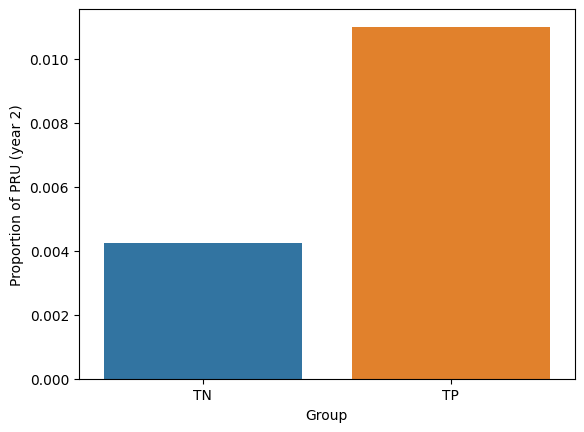}
%  \caption{ATTENDANCE\_YEAR4}
  \label{fig:sfig_s1_pru}
\end{subfigure}
\begin{subfigure}{.5\textwidth}
  \centering
  \includegraphics[width=1\linewidth]{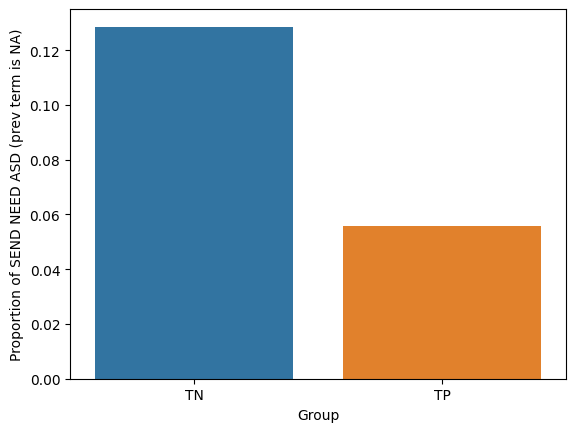}
%  \caption{AGE\_AT\_LOCALITY\_DECISION}
  \label{fig:sfig_s1_send_needs}
\end{subfigure}%
\begin{subfigure}{.5\textwidth}
  \centering
  \includegraphics[width=1\linewidth]{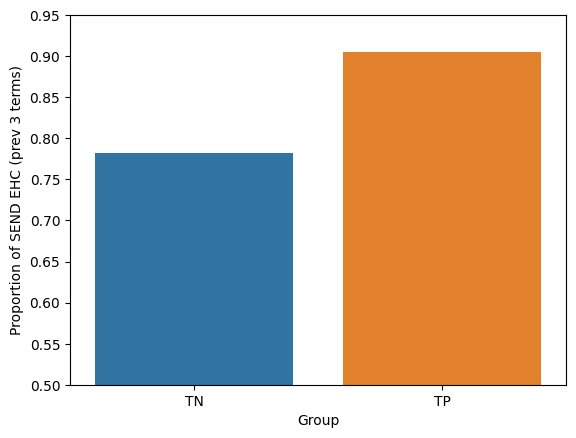}
%  \caption{ATTENDANCE\_YEAR4}
  \label{fig:sfig_s1_send_ehc}
\end{subfigure}
\captionsetup{labelformat=empty}
\caption{\textbf{Figure S1.} \textsc{eh support}: Profile of young people correctly classified as TN and TP. Features \textsc{age at locality decision}, \textsc{pru (year 2)}, \textsc{send need asd (prev term is na)} and \textsc{send ehc (prev 3 term)}.}
\label{fig:sfig_s1}
\end{figure}

\begin{figure}[htb]
\begin{subfigure}{.5\textwidth}
  \centering
  \includegraphics[width=1\linewidth]{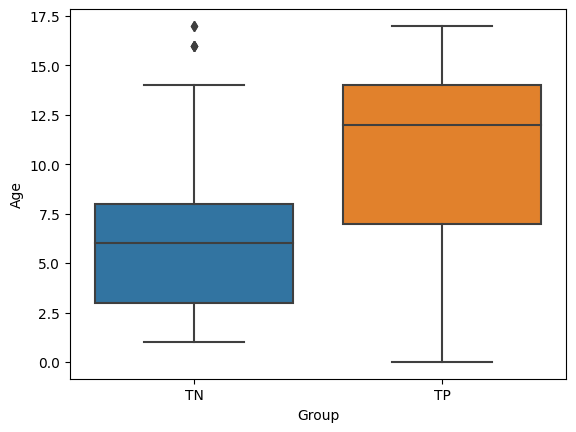}
%  \caption{AGE\_AT\_LOCALITY\_DECISION}
  \label{fig:sfig_s2_age}
\end{subfigure}%
\begin{subfigure}{.5\textwidth}
  \centering
  \includegraphics[width=1\linewidth]{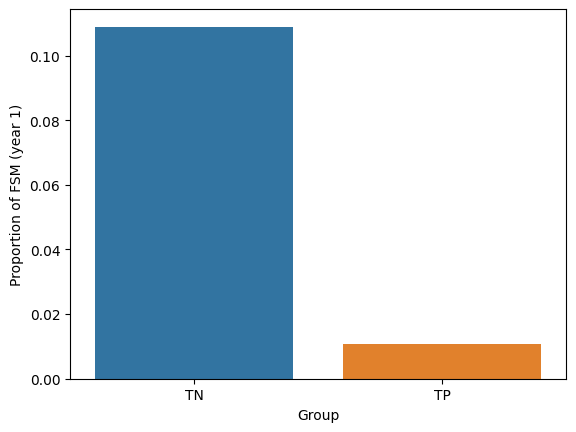}
%  \caption{ATTENDANCE\_YEAR4}
  \label{fig:sfig_s2_fsm}
\end{subfigure}
\begin{subfigure}{.5\textwidth}
  \centering
  \includegraphics[width=1\linewidth]{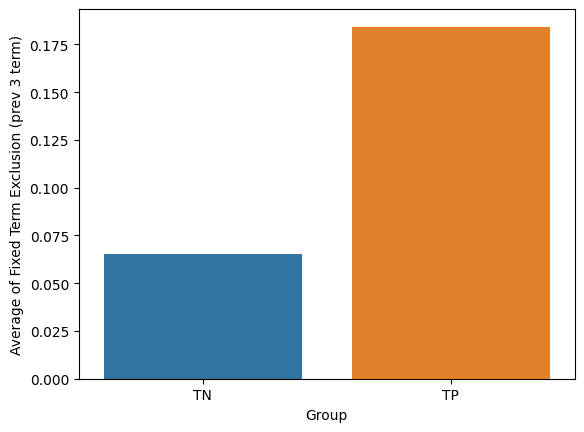}
%  \caption{AGE\_AT\_LOCALITY\_DECISION}
  \label{fig:sfig_s2_exc_fixed}
\end{subfigure}%
\begin{subfigure}{.5\textwidth}
  \centering
  \includegraphics[width=1\linewidth]{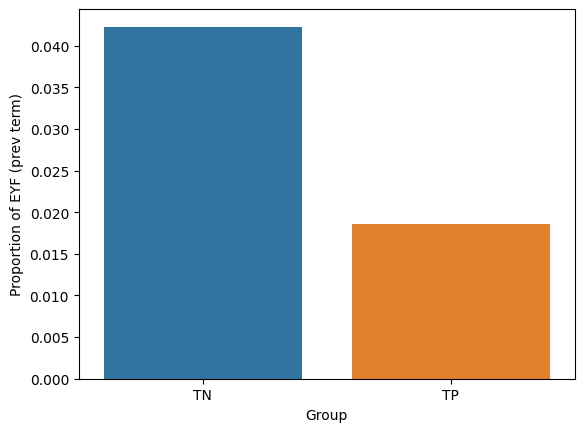}
%  \caption{ATTENDANCE\_YEAR4}
  \label{fig:sfig_s2_eyf}
\end{subfigure}
 \captionsetup{labelformat=empty}
\caption{\textbf{Figure S2.} \textsc{some action}: Profile of young people correctly classified as TN and TP. Features \textsc{age at locality decision}, \textsc{fsm (year 1)}, \textsc{exc fixed (prev 3 term)} and \textsc{eyf (prev term)}.}
\label{fig:fig_s2}
\end{figure}

\begin{figure}[htb]
\begin{subfigure}{.5\textwidth}
  \centering
  \includegraphics[width=1\linewidth]{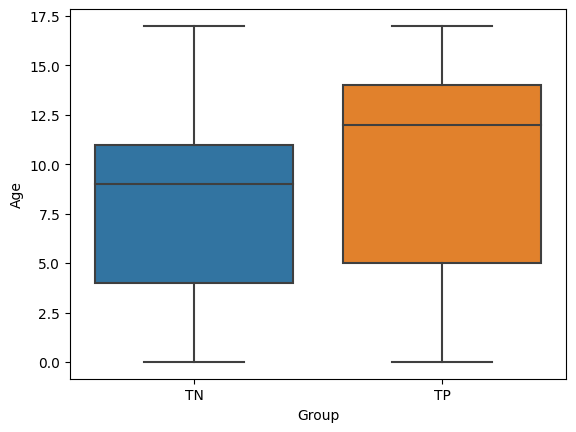}
  \label{fig:sfig_s3_age}
\end{subfigure}%
\begin{subfigure}{.5\textwidth}
  \centering
  \includegraphics[width=1\linewidth]{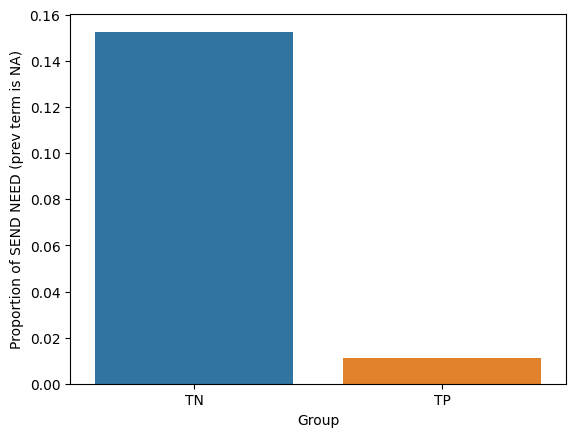}
  \label{fig:sfig_s3_send_need}
\end{subfigure}
 \captionsetup{labelformat=empty}
\caption{\textbf{Figure S3.} \textsc{no action}: Profile of young people correctly classified as TN and TP. Features \textsc{age at locality decision} and \textsc{send need (prev term is an)}.}
\label{fig:fig_s3}
\end{figure}

\begin{figure}[htb]
\begin{subfigure}{.5\textwidth}
  \centering
  \includegraphics[width=1\linewidth]{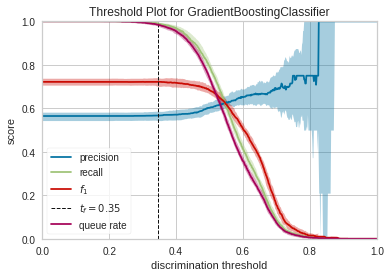}
  \label{fig:sfig_s4_gbc}
\end{subfigure}%
\begin{subfigure}{.5\textwidth}
  \centering
  \includegraphics[width=1\linewidth]{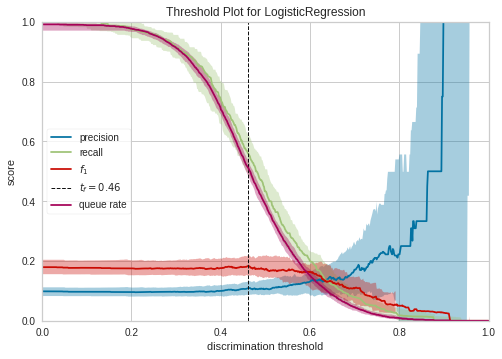}
  \label{fig:sfig_s4_lr2}
\end{subfigure}
 \captionsetup{labelformat=empty}
\caption{\textbf{Figure S4.} Threshold analysis for the GBC (\textsc{some action} model) and the LR classifier (\textsc{no action} model).}
\label{fig:fig_s4}
\end{figure}

\begin{figure}[htb]
\begin{subfigure}{.5\textwidth}
  \centering
  \includegraphics[width=1\linewidth]{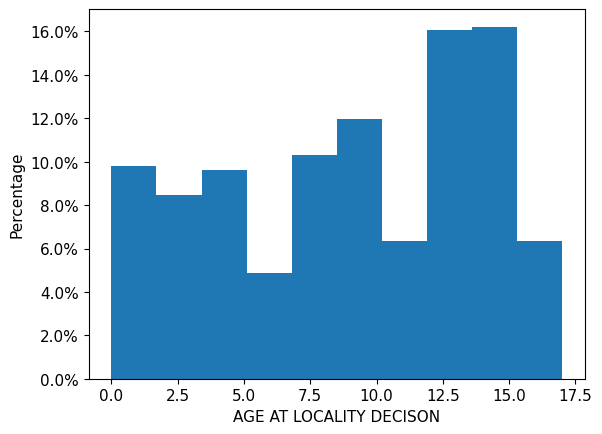}
%  \caption{AGE\_AT\_LOCALITY\_DECISION}
  \label{fig:sfig_s5_age}
\end{subfigure}%
\begin{subfigure}{.5\textwidth}
  \centering
  \includegraphics[width=1\linewidth]{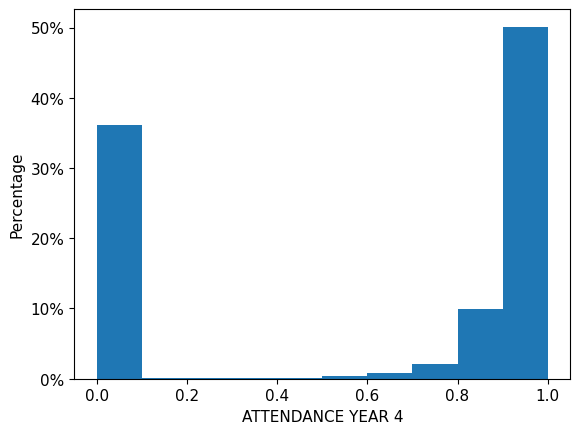}
%  \caption{ATTENDANCE\_YEAR4}
  \label{fig:sfig_s5_attendance_year4}
\end{subfigure}
\captionsetup{labelformat=empty}
\caption{\textbf{Figure S5.} Histogram for the features \textsc{age at locality decision} and \textsc{attendance (year 4)}.}
\label{fig:fig_s5}
\end{figure}

\begin{table}[htb]
    \centering
    \begin{tabular}{c|c|c|c}
    \hline
        Classifier & AUC & Recall & Precision \\ \hline
        Gradient Boosting  & 0.6127 & 0.3699 & 0.5184 \\ %\hline
        Logistic Regression & 0.6003 & 0.3743 & 0.5171 \\ %\hline
        Linear Discriminant  & 0.5976 & 0.3762 & 0.5145 \\ %\hline
        CatBoost & 0.5947 & 0.3792 & 0.5055 \\ %\hline
        AdaBoost  & 0.5848 & 0.3624 & 0.5269 \\ %\hline
        Random Forest & 0.5785 & 0.3663 & 0.4914 \\ %\hline
        Extra Trees & 0.5600 & 0.3678 & 0.4869 \\ %\hline
        Naive Bayes & 0.5472 & 0.3470 & 0.5073 \\ %\hline
        K Neighbors  & 0.5369 & 0.3579 & 0.4772 \\ %\hline
        Quadratic Discriminant & 0.5281 & 0.3287 & 0.4675 \\ %\hline
        Decision Tree  & 0.5254 & 0.3647 & 0.4719 \\ %\hline
        Dummy  & 0.5000 & 0.3333 & 0.3169 \\ %\hline
        Ridge  & 0.0000 & 0.3734 & 0.5187 \\ %\hline
        SVM (Linear Kernel) & 0.0000 & 0.3659 & 0.5084 \\ \hline
    \end{tabular}
\captionsetup{labelformat=empty}
     \caption{\label{tab7} \textbf{Table S6.} Multi-class models: Validation performance of the ML models. Single run stratified 10-fold CV.}
\end{table}

\begin{table}[htb]
    \centering 
    \begin{tabular}{c|c|c|c}
    \hline
        Classifier & AUC & Recall & Precision \\ \hline
        GB & 0.6114 $\pm$ 0.0075 & 0.3688 $\pm$ 0.0050 & 0.5168 $\pm$  0.0141	 \\ %\hline
%            & std & 0.0049 & 0.0075 & 0.0050 & 0.0141 \\ \hline
        LR (with norm L1) & 0.6043 $\pm$ 0.0063 & \textbf{0.3993 $\pm$ 0.0055} & 0.5184 $\pm$ 0.0057 	 \\ %\hline
%        (with norm L1)  & std & 0.0062 & 0.0063 & 0.0055 & 0.0057 \\ \hline
        LR & 0.6027 $\pm$ 0.0067 & 0.3725 $\pm$ 0.0033 & 0.5229 $\pm$ 0.0179 \\ \hline
%            & std & 0.0049 & 0.0067 & 0.0033 & 0.0179 \\ \hline
    \end{tabular}
    \captionsetup{labelformat=empty}
     \caption{\label{tab8} \textbf{Table S7.} Multi-class model: Validation performance of the ML models. Average and standard deviation of the metrics using stratified 10-fold CV across 30 iterations.} 
\end{table}

\begin{table}[htb]
    \centering 
    \begin{tabular}{c|c|c|c}
    \hline
        Classifier &  AUC &  Recall & Precision \\ \hline
        GBC & 0.6271 $\pm$ 0.0097 & \textbf{0.8133 $\pm$ 0.0180} & 0.3839 $\pm$ 0.0038 \\ %\hline
        LR & 0.6198 $\pm$ 0.0087  & \textbf{0.8242 $\pm$  0.0125} & 0.3781 $\pm$ 0.0035 \\ \hline
    \end{tabular}
    \captionsetup{labelformat=empty}
    \caption{\label{tab2b} \textbf{Table S8.} \textsc{eh support} model: Validation performance for the best models. Average and standard deviation of the performance metrics in the validation sets (stratified 10-fold CV) across 30 iterations. %\textcolor{red}{something seems to have gone wrong with the AUC values} \textcolor{red}{can you please include the optimal ROC points? are these R and P values the ORPs? }
    }
\end{table}

\begin{table}[htb]
    \centering 
    \begin{tabular}{c|c|c|c}
    \hline
        Classifier & AUC & Recall & Precision \\ \hline
%        LR & 0.5611 $\pm$ 0.0117 & 0.0035 $\pm$ 0.0030 & 0.5367 $\pm$ 0.4411	 \\ %\hline
        LR & 0.5590 $\pm$ 0.0124 & \textbf{0.5980 $\pm$ 0.0340} & 0.1132 $\pm$ 0.0045 \\ %\hline
        CB & 0.5763 $\pm$ 0.0123  & 0.0031 $\pm$ 0.0029 & 0.1664 $\pm$ 0.1713	 \\ %\hline
        RF & 0.5438 $\pm$ 0.0164 & 0.0239 $\pm$ 0.0068 & 0.1786 $\pm$  0.0532 	 \\ \hline
    \end{tabular}
    \captionsetup{labelformat=empty}
    \caption{\label{tab6b} \textbf{Table S9.} \textsc{no action} model: Validation performance of the ML models. Average and standard deviation of the metrics using stratified 10-fold CV across 30 iterations.} 
\end{table}

\begin{table}[htb]
    \centering 
    \begin{tabular}{c|c|c|c|c|c}
    \hline
       Model & Classifier & AUC & Recall & Precision & Optimal ROC point (FPR, TPR) \\ \hline
       \textsc{eh support}& LR & 0.6280 & 0.8269 & 0.3784 & (0.3401, 0.5311) \\ %\hline
       \textsc{some action}& GBC & 0.6002 & 0.7908 & 0.3071 & (0.3683, 0.5284) \\ %\hline
    \textsc{no action} & LR & 0.5598 & 0.6270 & 0.1175 & (0.5404, 0.6430) \\ \hline
    \end{tabular}
    \captionsetup{labelformat=empty}
     \caption{\label{tab8b} \textbf{Table S10.} Test performance of the best models.} 
\end{table}

\begin{landscape}
\begin{table}[htb]
    \centering
    {\footnotesize
    \begin{tabular}{c|c|c|c|c|c|c|c|c|c}
    \hline
        Feature & mean & std & \% missing & \% NA & Feature & mean & std & \% missing & \% NA \\ \hline
        AGE AT LOCALITY DECISION & 8.98 & 4.92 & 0.00 & 0.00 & EXC LUNCH PREV TERM PCT & 0.00 & 0.00 & 6.76 & 22.59 \\ %\hline
        LOCALITY DECISION INCIDENCE & 1.60 & 0.95 & 0.00 & 0.00 & EXC LUNCH PREV 3 TERMS PCT & 0.00 & 0.03 & 6.92 & 18.76 \\ %\hline
        ABS AUTH PREV TERM & 0.05 & 0.11 & 5.00 & 22.59 & EXC LUNCH YEAR 1 PCT & 0.00 & 0.01 & 9.98 & 16.82 \\ %\hline
        ABS AUTH PREV 3 TERMS & 0.05 & 0.08 & 6.68 & 18.75 & EXC LUNCH YEAR 2 PCT & 0.00 & 0.03 & 11.28 & 20.10 \\ %\hline
        ABS AUTH YEAR 1 & 0.05 & 0.09 & 9.98 & 16.82 & EXC LUNCH YEAR 3 PCT & 0.00 & 0.05 & 11.20 & 24.76 \\ %\hline
        ABS AUTH YEAR 2 & 0.04 & 0.07 & 7.62 & 19.99 & EXC LUNCH YEAR 4 PCT & 0.00 & 0.04 & 12.80 & 29.74 \\ %\hline
        ABS AUTH YEAR 3 & 0.06 & 0.09 & 6.64 & 24.74 & EXC LUNCH YEAR 5 PCT & 0.00 & 0.01 & 6.34 & 20.46 \\ %\hline
        ABS AUTH YEAR 4 & 0.06 & 0.08 & 6.50 & 29.46 & EXC FIXED PREV TERM CNT & 0.06 & 0.35 & 6.07 & 22.68 \\ %\hline
        ABS AUTH YEAR 5 & 0.05 & 0.07 & 8.41 & 34.13 & EXC FIXED PREV 3 TERMS CNT & 0.00 & 0.01 & 7.24 & 18.75 \\ %\hline
        ABS UNAUTH PREV TERM & 0.04 & 0.09 & 5.00 & 22.59 & EXC FIXED YEAR 1 CNT & 0.00 & 0.00 & 9.98 & 16.82 \\ %\hline
        ABS UNAUTH PREV 3 TERMS & 0.04 & 0.07 & 6.68 & 18.75 & EXC FIXED YEAR 2 CNT & 0.00 & 0.09 & 13.47 & 19.99 \\ %\hline
        ABS UNAUTH YEAR 1 & 0.05 & 0.08 & 9.98 & 16.82 & EXC FIXED YEAR 3 CNT & 0.00 & 0.00 & 12.43 & 24.74 \\ %\hline
        ABS UNAUTH YEAR 2 & 0.04 & 0.06 & 7.62 & 19.99 & EXC FIXED YEAR 4 CNT & 0.00 & 0.00 & 13.29 & 29.46 \\ %\hline
        ABS UNAUTH YEAR 3 & 0.03 & 0.08 & 6.64 & 24.74 & EXC FIXED YEAR 5 CNT & 0.00 & 0.00 & 14.52 & 34.13 \\ %\hline
        ABS UNAUTH YEAR 4 & 0.02 & 0.06 & 6.50 & 29.46 & EXC FIXED PREV TERM PCT & 0.00 & 0.00 & 6.76 & 22.59 \\ %\hline
        ABS UNAUTH YEAR 5 & 0.01 & 0.03 & 8.41 & 34.13 & EXC FIXED PREV 3 TERMS PCT & 0.20 & 0.85 & 6.92 & 18.76 \\ %\hline
        ATTENDANCE PREV TERM & 0.91 & 0.14 & 5.00 & 22.59 & EXC FIXED YEAR 1 PCT & 0.25 & 0.97 & 9.98 & 16.82 \\ %\hline
        ATTENDANCE PREV 3 TERMS & 0.92 & 0.11 & 6.68 & 18.75 & EXC FIXED YEAR 2 PCT & 0.20 & 0.92 & 11.28 & 20.10 \\ %\hline
        ATTENDANCE YEAR 1 & 0.90 & 0.13 & 9.98 & 16.82 & EXC FIXED YEAR 3 PCT & 0.14 & 0.78 & 11.20 & 24.76 \\ %\hline
        ATTENDANCE YEAR 2 & 0.92 & 0.09 & 7.62 & 19.99 & EXC FIXED YEAR 4 PCT & 0.08 & 0.52 & 12.80 & 29.74 \\ %\hline
        ATTENDANCE YEAR 3 & 0.91 & 0.13 & 6.64 & 24.74 & EXC FIXED YEAR 5 PCT & 0.25 & 0.97 & 6.34 & 20.46 \\ %\hline
        ATTENDANCE YEAR 4 & 0.93 & 0.10 & 6.50 & 29.46 & PRU PREV TERM & 0.01 & 0.09 & 1.31 & 14.31 \\ %\hline
        ATTENDANCE YEAR 5 & 0.94 & 0.08 & 8.41 & 34.13 & PRU PREV 3 TERMS & 0.01 & 0.09 & 0.61 & 12.52 \\ %\hline
        EXC PERMANENT PREV TERM CNT & 0.00 & 0.03 & 6.07 & 22.68 & PRU YEAR 1 & 0.01 & 0.10 & 6.27 & 8.68 \\ %\hline
        EXC PERMANENT PREV 3 TERMS CNT & 0.00 & 0.04 & 6.92 & 18.76 & PRU YEAR 2 & 0.01 & 0.09 & 3.02 & 11.52 \\ %\hline
        EXC PERMANENT YEAR 1 CNT & 0.00 & 0.03 & 9.98 & 16.82 & PRU YEAR 3 & 0.01 & 0.09 & 1.27 & 16.07 \\ %\hline
        EXC PERMANENT YEAR 2 CNT & 0.00 & 0.03 & 11.28 & 20.10 & PRU YEAR 4 & 0.01 & 0.08 & 1.01 & 20.68 \\ %\hline
        EXC PERMANENT YEAR 3 CNT & 0.00 & 0.05 & 11.20 & 24.76 & PRU YEAR 5 & 0.01 & 0.08 & 1.57 & 25.38 \\ %\hline
        EXC PERMANENT YEAR 4 CNT & 0.00 & 0.03 & 12.80 & 29.74 & HOME EDUCATED PREV TERM & 0.01 & 0.10 & 1.30 & 14.31 \\ %\hline
        EXC PERMANENT YEAR 5 CNT & 0.00 & 0.03 & 6.34 & 20.46 & HOME EDUCATED PREV 3 TERMS & 0.02 & 0.13 & 0.59 & 12.52 \\ %\hline
        EXC LUNCH PREV TERM CNT & 0.00 & 0.02 & 6.07 & 22.68 & HOME EDUCATED YEAR 1 & 0.02 & 0.15 & 6.26 & 8.68 \\ %\hline
        EXC LUNCH PREV 3 TERMS CNT  & 0.00 & 0.00 & 7.24 & 18.75 & HOME EDUCATED YEAR 2 & 0.02 & 0.14 & 3.02 & 11.52 \\ %\hline
        EXC LUNCH YEAR 1 CNT & 0.00 & 0.00 & 9.98 & 16.82 & HOME EDUCATED YEAR 3 & 0.01 & 0.11 & 1.27 & 16.07 \\ %\hline
        EXC LUNCH YEAR 2 CNT & 0.00 & 0.00 & 13.47 & 19.99 & HOME EDUCATED YEAR 4 & 0.01 & 0.10 & 1.01 & 20.68 \\ %\hline
        EXC LUNCH YEAR 3 CNT & 0.00 & 0.00 & 12.43 & 24.74 & HOME EDUCATED YEAR 5 & 0.02 & 0.15 & 1.55 & 25.33 \\ %\hline
        EXC LUNCH YEAR 4 CNT & 0.00 & 0.00 & 13.29 & 29.46 & MISSING EDUCATION PREV TERM & 0.02 & 0.13 & 1.30 & 14.31 \\ %\hline
        EXC LUNCH YEAR 5 CNT & 0.00 & 0.00 & 14.52 & 34.13 & MISSING EDUCATION PREV 3 TERMS & 0.03 & 0.17 & 0.58 & 12.52 \\ %\hline
        IDACI & 0.13 & 0.09 & 0.32 & 0.00 &  &  &  &  &  \\ \hline
    \end{tabular}
    }
   \captionsetup{labelformat=empty}
     \caption{\label{tab_descr_stat} \textbf{Table S11.} Descriptive statistics: mean, standard deviation (std), percentage of missing values and percentage of not applicable (NA) cells by feature.} 
\end{table}

\begin{table}[htb]
    \centering
    {\footnotesize
    \begin{tabular}{c|c|c|c|c|c|c|c|c|c}
    \hline
        Feature & mean & std & \% missing & \% NA & Feature & mean & std & \% missing & \% NA \\ \hline
        MISSING EDUCATION PREV TERM & 0.02 & 0.13 & 1.30 & 14.31 & SEND EHC YEAR 2 & 0.09 & 0.29 & 7.21 & 3.11 \\ %\hline
        MISSING EDUCATION PREV 3 TERMS & 0.03 & 0.17 & 0.58 & 12.52 & SEND EHC YEAR 3 & 0.08 & 0.28 & 7.42 & 6.16 \\ %\hline
        MISSING EDUCATION YEAR 1 & 0.04 & 0.19 & 6.24 & 8.66 & SEND EHC YEAR 4 & 0.07 & 0.26 & 7.92 & 9.76 \\ %\hline
        MISSING EDUCATION YEAR 2 & 0.04 & 0.19 & 2.95 & 11.52 & SEND EHC YEAR 5 & 0.06 & 0.24 & 8.39 & 13.47 \\ %\hline
        MISSING EDUCATION YEAR 3 & 0.02 & 0.14 & 1.23 & 16.07 & SEND NEED ASD PREV TERM & 0.02 & 0.14 & 7.74 & 7.72 \\ %\hline
        MISSING EDUCATION YEAR 4 & 0.01 & 0.07 & 1.00 & 20.68 & SEND NEED HI PREV TERM & 0.00 & 0.00 & 7.85 & 7.77 \\ %\hline
        MISSING EDUCATION YEAR 5 & 0.00 & 0.04 & 1.57 & 25.38 & SEND NEED MLD PREV TERM & 0.00 & 0.00 & 7.85 & 7.77 \\ %\hline
        TRANSFER PHASED PREV TERM & 0.05 & 0.21 & 1.28 & 14.26 & SEND NEED MSI PREV TERM & 0.00 & 0.00 & 7.85 & 7.77 \\ %\hline
        TRANSFER PHASED PREV 3 TERMS & 0.14 & 0.34 & 0.59 & 12.47 & SEND NEED PD PREV TERM & 0.00 & 0.00 & 7.85 & 7.77 \\ %\hline
        TRANSFER PHASED YEAR 1 & 0.13 & 0.33 & 6.25 & 8.48 & SEND NEED PMLD PREV TERM & 0.00 & 0.00 & 7.85 & 7.77 \\ %\hline
        TRANSFER PHASED YEAR 2 & 0.15 & 0.36 & 2.96 & 11.34 & SEND NEED SEMH PREV TERM & 0.00 & 0.00 & 7.85 & 7.77 \\ %\hline
        TRANSFER PHASED YEAR 3 & 0.17 & 0.37 & 1.25 & 15.90 & SEND NEED SLCN PREV TERM & 0.00 & 0.00 & 7.85 & 7.77 \\ %\hline
        TRANSFER PHASED YEAR 4 & 0.20 & 0.40 & 0.99 & 20.44 & SEND NEED SLD PREV TERM & 0.00 & 0.00 & 7.85 & 7.77 \\ %\hline
        TRANSFER PHASED YEAR 5 & 0.22 & 0.41 & 1.52 & 25.24 & SEND NEED SPLD PREV TERM & 0.00 & 0.00 & 7.85 & 7.77 \\ %\hline
        TRANSFER MIDTERM PREV TERM & 0.00 & 0.00 & 1.31 & 14.31 & SEND NEED VI PREV TERM & 0.00 & 0.00 & 7.85 & 7.77 \\ %\hline
        TRANSFER MIDTERM PREV 3 TERMS & 0.00 & 0.00 & 0.61 & 12.52 & FSM PREV TERM & 0.02 & 0.14 & 1.31 & 14.30 \\ %\hline
        TRANSFER MIDTERM YEAR 1 & 0.00 & 0.00 & 6.27 & 8.68 & FSM PREV 3 TERMS & 0.06 & 0.24 & 0.61 & 12.51 \\ %\hline
        TRANSFER MIDTERM YEAR 2 & 0.00 & 0.00 & 3.02 & 11.52 & FSM YEAR 1 & 0.05 & 0.22 & 6.25 & 8.68 \\ %\hline
        TRANSFER MIDTERM YEAR 3 & 0.00 & 0.00 & 1.27 & 16.07 & FSM YEAR 2 & 0.06 & 0.24 & 3.02 & 11.52 \\ %\hline
        TRANSFER MIDTERM YEAR 4 & 0.00 & 0.00 & 1.01 & 20.68 & FSM YEAR 3 & 0.06 & 0.24 & 1.25 & 16.06 \\ %\hline
        TRANSFER MIDTERM YEAR 5 & 0.00 & 0.00 & 1.57 & 25.38 & FSM YEAR 4 & 0.07 & 0.25 & 0.95 & 20.68 \\ %\hline
        SEND REFERRAL PREV TERM & 0.01 & 0.07 & 1.66 & 2.11 & FSM YEAR 5 & 0.07 & 0.26 & 1.56 & 25.38 \\ %\hline
        SEND REFERRAL PREV 3 TERMS & 0.02 & 0.13 & 1.83 & 1.60 & NEET PREV TERM & 0.02 & 0.15 & 0.81 & 94.56 \\ %\hline
        SEND REFERRAL YEAR 1 & 0.01 & 0.12 & 2.49 & 0.59 & NEET PREV 3 TERMS & 0.04 & 0.21 & 3.93 & 89.86 \\ %\hline
        SEND REFERRAL YEAR 2 & 0.02 & 0.14 & 2.10 & 1.36 & NEET YEAR 1 & 0.05 & 0.23 & 2.26 & 80.57 \\ %\hline
        SEND REFERRAL YEAR 3 & 0.01 & 0.12 & 1.83 & 2.27 & NEET YEAR 2 & 0.06 & 0.24 & 0.93 & 88.59 \\ %\hline
        SEND REFERRAL YEAR 4 & 0.02 & 0.13 & 1.80 & 3.15 & NEET YEAR 3 & 0.04 & 0.21 & 0.25 & 95.00 \\ %\hline
        SEND REFERRAL YEAR 5 & 0.01 & 0.11 & 1.75 & 4.04 & NEET YEAR 4 & 0.01 & 0.11 & 0.07 & 98.18 \\ %\hline
        SEND SUPPORT PREV TERM & 0.17 & 0.38 & 5.61 & 6.24 & NEET YEAR 5 & 0.00 & 0.00 & 0.01 & 99.70 \\ %\hline
        SEND SUPPORT PREV 3 TERMS & 0.18 & 0.38 & 6.18 & 4.09 & EYF PREV TERM & 1.00 & 0.00 & 2.92 & 95.08 \\ %\hline
        SEND SUPPORT YEAR 1 & 0.18 & 0.38 & 8.05 & 0.96 & EYF PREV 3 TERMS & 1.00 & 0.00 & 2.43 & 93.85 \\ %\hline
        SEND SUPPORT YEAR 2 & 0.17 & 0.38 & 7.21 & 3.11 & EYF EVER & 0.49 & 0.50 & 0.00 & 59.17 \\ %\hline
        SEND SUPPORT YEAR 3 & 0.18 & 0.38 & 7.42 & 6.16 & MALE & 0.54 & 0.50 & 0.00 & 0.00 \\ %\hline
        SEND SUPPORT YEAR 4 & 0.18 & 0.38 & 7.92 & 9.76 & FEMALE & 0.45 & 0.50 & 0.00 & 0.00 \\ %\hline
        SEND SUPPORT YEAR 5 & 0.18 & 0.38 & 8.39 & 13.47 & OTHER & 0.00 & 0.07 & 0.00 & 0.00 \\ %\hline
        SEND EHC PREV TERM & 0.09 & 0.28 & 5.61 & 6.24 & SOME ACTION & 0.57 & 0.50 & 0.00 & 0.00 \\ %\hline
        SEND EHC PREV 3 TERMS & 0.09 & 0.29 & 6.18 & 4.09 & NO ACTION & 0.10 & 0.30 & 0.00 & 0.00 \\ %\hline
        SEND EHC YEAR 1 & 0.10 & 0.30 & 8.05 & 0.96 & EH SUPPORT & 0.33 & 0.47 & 0.00 & 0.00 \\ \hline
    \end{tabular}
    }
    \captionsetup{labelformat=empty}
    \caption{\label{tab_descr_stat_cont} \textbf{Table S11 continued.} Descriptive statistics: mean, standard deviation (std), percentage of missing values and percentage of not applicable (NA) cells by feature.} 
\end{table}

\begin{table}[htb]
    \centering
    {\footnotesize
    \begin{tabular}{c|c|c|c|c|c|c|c|c}
    \hline
       Model & Feature & Group test & $\hat{p}_1$ & $n_1$ & $\hat{p}_2$ & $n_2$ & Test Statistic $|Z|$ & p-value \\ \hline
       \multirow{11}{*}{\textsc{eh support}} & \textsc{gender}  & \textsc{male} vs \textsc{female} & 0.155 & 846 & 0.191 & 577 & 1.749 & 0.080 \\ \cline{2-9}
            & \multirow{10}{*}{\textsc{idaci}}  & \textsc{idaci 1} vs \textsc{idaci 2} & 0.188 & 287 & 0.216 & 310 & 0.853 & 0.394 \\ 
            &  & \textsc{idaci 1} vs \textsc{idaci 3} & 0.188 & 287 & 0.129 & 256 & 1.893 & 0.058 \\
            &  & \textsc{idaci 1} vs \textsc{idaci 4} & 0.188 & 287 & 0.168 & 268 & 0.616 & 0.538 \\
            &  & \textsc{idaci 1} vs \textsc{idaci 5} & 0.188 & 287 & 0.159 & 302 & 0.929 & 0.353 \\
            &  & \textsc{idaci 2} vs \textsc{idaci 3} & 0.216 & 310 & 0.129 & 256 & 2.771 & \textbf{0.006} \\
            &  & \textsc{idaci 2} vs \textsc{idaci 4} & 0.216 & 310 & 0.168 & 268 & 1.468 & 0.142 \\
            &  & \textsc{idaci 2} vs \textsc{idaci 5} & 0.216 & 310 & 0.159 & 302 & 1.812 & 0.070 \\ 
            &  & \textsc{idaci 3} vs \textsc{idaci 4} & 0.129 & 256 & 0.168 & 268 & 1.258 & 0.208 \\
            &  & \textsc{idaci 3} vs \textsc{idaci 5} & 0.129 & 256 & 0.159 & 302 & 1.010 & 0.312 \\
            &  & \textsc{idaci 4} vs \textsc{idaci 5} & 0.168 & 268 & 0.159 & 302 & 0.289 & 0.772 \\ \hline
    \multirow{15}{*}{\textsc{some action}} & \textsc{gender}  & \textsc{male} vs \textsc{female} & 0.200 & 1309 & 0.216 & 1121 & 0.967 & 0.333 \\ \cline{2-9}
        & \textsc{attendance (year 4)}  & $\leq 0.5$ vs $> 0.5$ & 0.362 & 873 & 0.124 & 1570 & 13.027 & \textbf{< 0.001} \\ \cline{2-9}
        & \multirow{3}{*}{\textsc{age class}}  & \textsc{below 7.5} vs \textsc{between 7.5 and 12.5} & 0.377 & 872 & 0.212 & 720 & 7.368 & \textbf{< 0.001} \\ 
            &  & \textsc{below 7.5} vs \textsc{above 12.5} & 0.377 & 872 & 0.034 & 851 & 19.546 & \textbf{< 0.001}\\
            &  & \textsc{between 7.5 and 12.5} vs \textsc{above 12.5} & 0.212 & 720 & 0.034 & 851 & 10.820 & \textbf{< 0.001}\\ \cline{2-9}       
        & \multirow{10}{*}{\textsc{idaci}}  & \textsc{idaci 1} vs \textsc{idaci 2} & 0.218 & 486 & 0.248 & 464 & 1.093 & 0.274 \\ 
            &  & \textsc{idaci 1} vs \textsc{idaci 3} & 0.218 & 486 & 0.212 & 542 & 0.233 & 0.815 \\
            &  & \textsc{idaci 1} vs \textsc{idaci 4} & 0.218 & 486 & 0.188 & 479 & 1.159 & 0.246 \\
            &  & \textsc{idaci 1} vs \textsc{idaci 5} & 0.218 & 486 & 0.180 & 472 & 1.475 & 0.140 \\
            &  & \textsc{idaci 2} vs \textsc{idaci 3} & 0.248 & 464 & 0.212 & 542 & 1.350 & 0.177 \\
            &  & \textsc{idaci 2} vs \textsc{idaci 4} & 0.248 & 464 & 0.188 & 479 & 2.234 & \textbf{0.025} \\
            &  & \textsc{idaci 2} vs \textsc{idaci 5} & 0.248 & 464 & 0.180 & 472 & 2.543 & \textbf{0.011} \\ 
            &  & \textsc{idaci 3} vs \textsc{idaci 4} & 0.212 & 542 & 0.188 & 479 & 0.958 & 0.338 \\
            &  & \textsc{idaci 3} vs \textsc{idaci 5} & 0.212 & 542 & 0.180 & 472 & 1.284 & 0.199 \\
            &  & \textsc{idaci 4} vs \textsc{idaci 5} & 0.188 & 479 & 0.180 & 472 & 0.318 & 0.750 \\ \hline
     \multirow{14}{*}{\textsc{no action}} & \textsc{gender}  & \textsc{male} vs \textsc{female} & 0.386 & 241 & 0.359 & 195 & 0.580 & 0.562 \\ \cline{2-9}
        & \multirow{3}{*}{\textsc{age class}}  & \textsc{below 7.5} vs \textsc{between 7.5 and 12.5} & 0.362 & 163 & 0.561 & 123 & 3.403 & \textbf{< 0.001} \\ 
            &  & \textsc{below 7.5} vs \textsc{above 12.5} & 0.362 & 163 & 0.232 & 151 & 2.551 & \textbf{0.010}\\
            &  & \textsc{between 7.5 and 12.5} vs \textsc{above 12.5} & 0.561 & 123 & 0.232 & 151 & 5.832 & \textbf{< 0.001}\\ \cline{2-9}       
        & \multirow{10}{*}{\textsc{idaci}}  & \textsc{idaci 1} vs \textsc{idaci 2} & 0.376 & 93 & 0.300 & 70 & 1.022 & 0.306 \\ 
            &  & \textsc{idaci 1} vs \textsc{idaci 3} & 0.376 & 93 & 0.406 & 89 & 0.414 & 0.679 \\
            &  & \textsc{idaci 1} vs \textsc{idaci 4} & 0.376 & 93 & 0.349 & 86 & 0.375 & 0.707 \\
            &  & \textsc{idaci 1} vs \textsc{idaci 5} & 0.376 & 93 & 0.414 & 94 & 0.532 & 0.595 \\
            &  & \textsc{idaci 2} vs \textsc{idaci 3} & 0.300 & 70 & 0.406 & 89 & 1.401 & 0.161 \\
            &  & \textsc{idaci 2} vs \textsc{idaci 4} & 0.300 & 70 & 0.349 & 86 & 0.652 & 0.514 \\
            &  & \textsc{idaci 2} vs \textsc{idaci 5} & 0.300 & 70 & 0.414 & 94 & 1.526 & 0.127 \\ 
            &  & \textsc{idaci 3} vs \textsc{idaci 4} & 0.406 & 89 & 0.349 & 86 & 0.778 & 0.436 \\
            &  & \textsc{idaci 3} vs \textsc{idaci 5} & 0.406 & 89 & 0.414 & 94 & 0.109 & 0.912 \\
            &  & \textsc{idaci 4} vs \textsc{idaci 5} & 0.349 & 86 & 0.414 & 94 & 0.899 & 0.368 \\ \hline
    \end{tabular}
    }
    \captionsetup{labelformat=empty}
    \caption{\label{tab_stat_test} \textbf{Table S12.} Two-sample Z-test for proportions. Comparison between the false negative rates ($\hat{p}$) and sensitive features. Sample size ($n$), test statistic Z and p-value. Values in bold reject the null hypothesis at significance level $\alpha = 0.05$.} 
\end{table}
\end{landscape}

\begin{table}[htb]
\centering
    {\footnotesize
    \begin{tabular}{c|cc} \hline
    \textsc{ethnicity} & Entire dataset & Test set \\ \hline
    White  & 11\,053 & 3\,363  \\
    Asian  & 588 & 170  \\
    Black  & 154 & 58  \\
    Mixed  & 456 & 131  \\
    Other  & 148 & 46  \\
    Not stated  & 631 & 171  \\
    Missing & 1\,330 & 369 \\ \hline
    \end{tabular}
    }
    \captionsetup{labelformat=empty}
    \caption{\label{tab_ethinicity_values} \textbf{Table S13.} Frequency distribution of \textsc{ethnicity} in the entire dataset and in the test set.} 
\end{table}

\begin{table}[htb]
    \centering
    {\footnotesize
    \begin{tabular}{c|c|c|c}
    \hline
       Model & \textsc{ethnicity} & Correctly classified (\%) & Misclassified (\%)\\ \hline
       \multirow{2}{*}{\textsc{eh support}} & \textsc{white} & 48.65 & 51.35 \\ 
            &  \textsc{not white} & 51.53 & 48.47 \\ \hline
       
       \multirow{2}{*}{\textsc{some action}} & \textsc{white} & 57.89 & 42.11 \\ 
            &  \textsc{not white} & 57.38 & 42.62\\ \hline

        \multirow{2}{*}{\textsc{no action}} & \textsc{white} & 48.41 & 51.59 \\ 
            &  \textsc{not white} & 47.54 & 52.46 \\ \hline
    
    \end{tabular}
    }
    \captionsetup{labelformat=empty}
    \caption{\label{tab_ethinicity} \textbf{Table S14.} Percentage of young people correctly classified and misclassified by the model in the test set. Category \textsc{not white} refers to young people who are not in the category \textsc{white} in Supplementary Table S14.} 
\end{table}

\bibliographystyle{unsrt}
\bibliography{references}  %%% Uncomment this line and comment out the ``thebibliography'' section below to use the external .bib file (using bibtex) .





% \end{thebibliography}

\end{document}